\documentclass[10pt,twocolumn,letterpaper]{article}

\usepackage{iccv}
\usepackage{times}
\usepackage{epsfig}
\usepackage{graphicx}
\usepackage{amsmath}
\usepackage{amssymb}

\usepackage{bm}
\usepackage{subfigure}
\usepackage{booktabs}
\usepackage{multirow}
\usepackage{bbding}
\usepackage{comment}
\usepackage{wrapfig}
\usepackage[table]{xcolor}

\newcommand*{\Perm}[2]{{}\!P^{#1}_{#2}}%

\usepackage[pagebackref=true,breaklinks=true,letterpaper=true,colorlinks,bookmarks=false]{hyperref}

\iccvfinalcopy 

\def\iccvPaperID{****} 

\ificcvfinal\fi

\begin{document}

\title{Optimization for Arbitrary-Oriented Object Detection via Representation Invariance Loss}

\author{Qi Ming$^1$,  Zhiqiang Zhou$^{1}$\thanks{Corresponding author},  Lingjuan Miao$^1$, Xue Yang$^2$,  Yunpeng Dong$^1$\\
$^1$ School of Automation, Beijing Institute of Technology\\
$^2$ Department of Computer Science and Engineering, Shanghai Jiao Tong University\\
{\tt\small chaser.ming@gmail.com}
}

\maketitle

\ificcvfinal\thispagestyle{empty}\fi

\begin{abstract}
Arbitrary-oriented objects exist widely in natural scenes, and thus the oriented object detection has received extensive attention in recent years. The mainstream rotation detectors use oriented bounding boxes (OBB) or quadrilateral bounding boxes (QBB) to represent the rotating objects. However, these methods suffer from the representation ambiguity for oriented object definition, which leads to suboptimal regression optimization and the inconsistency between the loss metric and the localization accuracy of the predictions. In this paper, we propose a Representation Invariance Loss (RIL) to optimize the bounding box regression for the rotating objects. Specifically, RIL treats multiple representations of an oriented object as multiple equivalent local minima, and hence transforms bounding box regression into an adaptive  matching process with these local minima. Then, the Hungarian matching algorithm is adopted to obtain the optimal regression strategy. We also propose a normalized rotation loss to alleviate the weak correlation between different variables and their unbalanced loss contribution in OBB representation. Extensive experiments on remote sensing datasets and scene text datasets show that our method achieves consistent and substantial improvement. The source code and trained models are available at https://github.com/ming71/RIDet.
\end{abstract}


\section{Introduction}
Arbitrary-oriented object detection has a wide range of application scenarios, such as scene text detection \cite{ma2018arbitrary,liao2018textboxes++,he2017single}, face detection \cite{shi2018real}, object detection in remote sensing images \cite{yang2018automatic,jiao2018densely,yang2018position,xia2018dota,yang2019scrdet,ding2019learning,yang2020scrdet++}, and 3D object detection \cite{zheng2020rotation}. In recent years, with the breakthroughs made by convolutional neural networks (CNNs) in the field of object detection  \cite{girshick2014rich,ren2016faster,liu2016ssd,redmon2016you}, a series of CNN-based rotation detectors have been proposed to achieve high-performance oriented object detection \cite{yang2020arbitrary,ming2020dynamic,ming2021cfc,liu2019omnidirectional,xu2020gliding,ming2021sparse}.

Unlike generic object detection that uses the horizontal bounding box to represent the objects, rotation detectors usually adopt the oriented bounding box (OBB) \cite{ding2019learning,ming2020dynamic,ming2021cfc} or the quadrilateral bounding box (QBB) \cite{xu2020gliding,liao2018textboxes++,liu2019omnidirectional} to describe the rotating objects, which induces the representation ambiguity. Whether using OBB or QBB, representation ambiguity indicates that an object can be represented in many different forms. These ambiguous representations constitute the representation space of a ground-truth (GT) object $g$, denoted as $\Omega(g)=\{g_0, g_1, g_2, ... \}$. Ideally, all representations in $\Omega(g)$ are equivalent local optimal solutions in the regression optimization process. However, the loss function of the current rotation detectors can only converge to the given GT representation $g_0$, and the rest $g_k \in \Omega(g)$ will cause a sharp increase in regression loss. The incorrect loss metric cannot truly reflect the localozation quality of the predictions, which makes the regression process hard to converge and further degrades the detection performance. We next separately analyze the influence of representation ambiguity on the two forms of bounding box definitions.  

\begin{figure}[t]
\begin{center}
\includegraphics[width=1.0\linewidth]{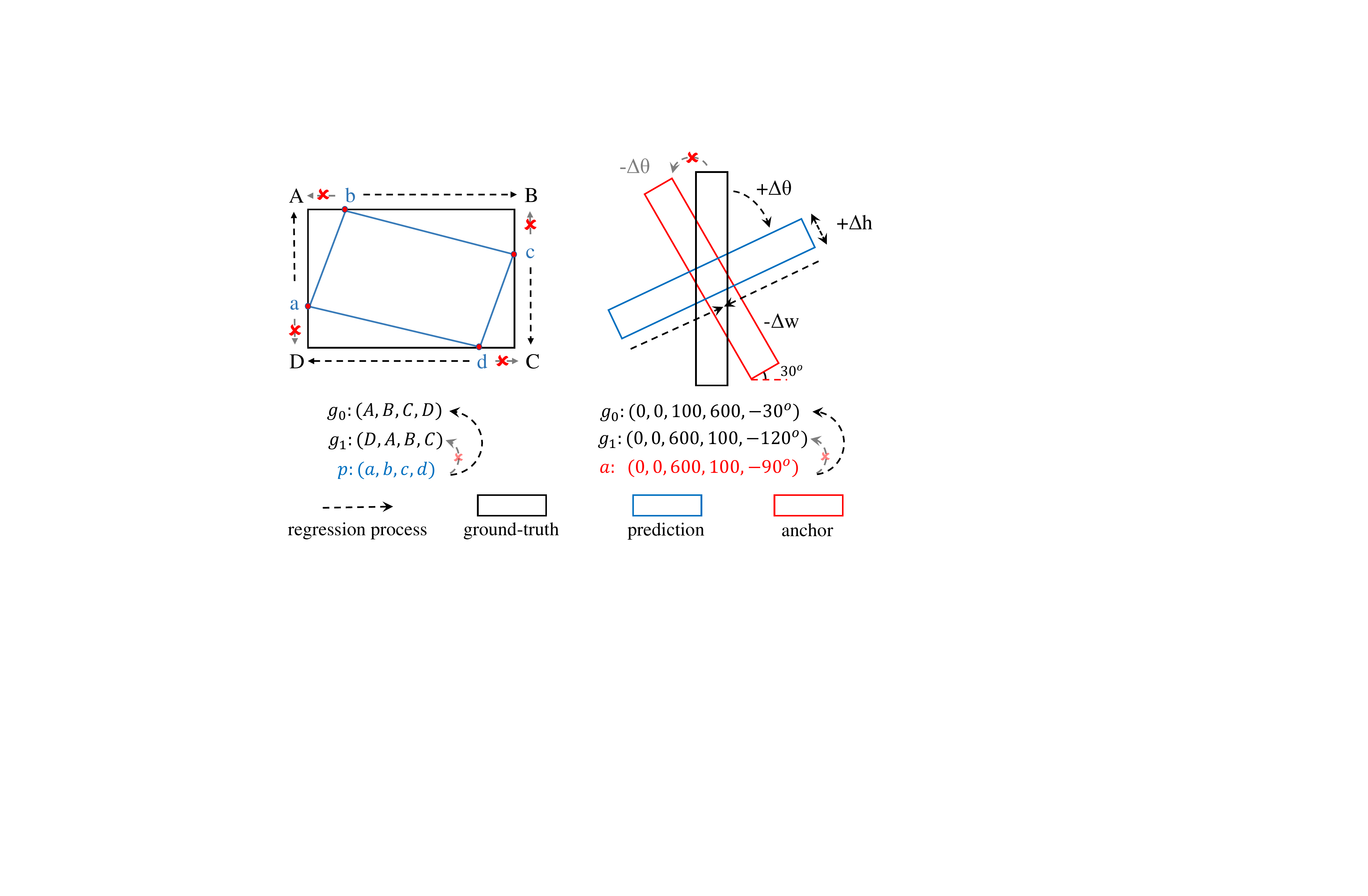}
\end{center}
   \caption{Examples of suboptimal regression process caused by the representation ambiguity under QBB (left) and OBB (right). $g_0$ is the given GT representation, and $g_1$ is one of the ambiguous representations of GT.}
\label{fig1}
\end{figure}

QBB directly uses four vertices of the quadrilateral to represent the oriented objects. The unordered vertices have $\Perm{4}{4}=24$ permutations, so theoretically there are 24 quadrilateral representations for one box in the representation space $\Omega$. These diverse representations provide more easily searchable local minima for regression optimization theoretically, but the existing methods do not consider all feasible instances, only to learn a given fixed-sequence quadrilateral representation. In this case, ambiguous representations are not effectively utilized but causes additional learning costs.  As shown in the left side of Fig.~\ref{fig1}, the predicted quadrilateral is close to the ground-truth box, and the ideal regression strategy is $\{(a\rightarrow D),(b\rightarrow A),(c\rightarrow B),(d\rightarrow C)\}$. But most current detectors will follow the established sequence for regression: $\{(a\rightarrow A),(b\rightarrow B),(c\rightarrow C),(d\rightarrow D)\}$. The suboptimal regression strategy would lead to a sharp loss boost, and thereby increasing the difficulty of network convergence.

Due to boundary problems \cite{yang2019scrdet,yang2020arbitrary} of angle definition of the oriented rectangle box, we suggest that the representation ambiguity still exists in OBB. On the one hand, the periodicity of angle brings a lot of equivalent representations. On the other hand, the interchangeability between width and height also derives many representations whose angles exceed the boundary. The current detectors cannot handle these ambiguous representations with out-of-bounds angles well. For example, the GT box in the right of Fig.~\ref{fig1} is given as $g_0: (0,0,100,600,-30^\circ)$. It can also be represented as $g_1: (0,0,600,100,-120^\circ)$ if without restriction of the angle boundary. For the preset anchor $a: (0,0,600,100,-90^\circ)$, the ideal optimization target is  $g_1$, which only requires a small counterclockwise angle offset. However, due to angle limitation in OBB, it can be only regressed to the suboptimal representation $g_0$, which would make the deviation of width, height, and angle very large, and the regression loss hard to converge. Besides, the contribution of different variables to the regression accuracy is unequal in OBB, and thus careful weighting is required to balance the loss terms of different variables.


Multiple representations due to representation ambiguity are essentially not bad for the regression task, because they would produce multiple equivalent local minima, and falling into any one of them would reach equivalently good convergence status, which would make the optimization process more flexible, and alleviate the problem of sharp boost and discontinuity  in regression loss. Current work, unfortunately, has not fully realized the above benefits and the rotation detectors usually suffer from suboptimal regression strategies that affect the network convergence and degrade detection performance.

In this paper, we propose a Representation Invariance Loss (RIL) that treats the bounding box regression as an adaptive  matching process with the local minima of multiple representations, and then searches for the optimal regression strategy dynamically through the Hungarian algorithm during training.  Specifically, for QBB, RIL transforms the point regression problem into a dynamic sequential-free vertices matching process, and then Hungarian loss is used to improve convergence and achieve superior performance. For OBB, the RIL allows the predictions with the out-of-bounds angles and adaptively searches for the optimal regression strategy from all ambiguous representations. Besides, a normalized rotation loss is adopted to alleviate the inconsistency between different variables.  
 
Rotation detectors trained with RIL treat the ambiguous representations as the equivalent local minima of the regression loss. These local minima provide better convergence than the unique minima defined by the GT representation. Therefore, during the training process, the regressor can dynamically select the optimal local minima in the representation space as the regression target according to the loss function. Based on the proposed RIL, we further build two Representation Invariant Detectors (RIDet) : RIDet-O and RIDet-Q, to prove the superiority of the proposed method. In summary, the main contribution is as follows:

1) We point out representation ambiguity in arbitrary-oriented object detection and analyzed the problems caused by multiple representations from the perspective of regression optimization.

2) The novel representation invariance loss (RIL) is proposed to transforms the bounding box regression task into an optimal matching process. RIL not only solves the inconsistency between the loss and localization quality caused by ambiguous representations, but also utilizes these local minima to improve network convergence and achieve better performance.

3) RIL can be easily applied to existing methods without any additional overhead. Extensive experiments on multiple rotation detectors and datasets prove the superiority of our approach.


\section{Related Work}

\subsection{Arbitrary-Oriented Object Detection}
Arbitrary-oriented objects appear widely in natural scenes, so oriented object detection has gradually become a hot topic in computer vision. In recent years, many advanced frameworks have been proposed to achieve accurate rotating object detection \cite{ming2020dynamic,liu2019omnidirectional,yang2021rethinking,han2021align,yang2019r3det,xu2020gliding}. 

Unlike generic object detection that uses the horizontal bounding box to represent objects, OBB and QBB are usually adopted for oriented object detection. For example, RoI Transformer \cite{ding2019learning} learns a rotated region of the horizontal region of interest to improve detection performance. RRPN \cite{ma2018arbitrary} generates inclined proposals with text orientation information to fit into the text region more accurately. R$^3$Det \cite{yang2019r3det}, S$^2$A-Net \cite{han2021align}, and CFC-Net \cite{ming2021cfc} obtain oriented anchors from horizontal anchors  to achieve better spatial alignment with oriented objects. TextBoxes++ \cite{liao2018textboxes++} predicts the vertex offsets from predefined anchor boxes to detect arbitrary-oriented text.   Xu \textit{et~al.} \cite{xu2020gliding} proposed a novel framework to regress four ratios on each corresponding side to detect multi-orientation objects. 

There are also some other arbitrary-oriented object representation methods, such as  polar coordinates \cite{zhou2020objects},  middle lines \cite{wei2020oriented}, ellipse representation \cite{liu2020efn}, and box boundary-aware vectors \cite{yi2020oriented}. Nevertheless, these methods are relatively complex and not general, so the mainstream methods are still OBB and QBB.

\subsection{Representation Problems of Oriented Objects}

Inappropriate representation of rotating objects will give rise to many problems. Some previous work summarized the issues as the boundary problem \cite{yang2019scrdet, yang2020arbitrary} in OBB representation, and suggested that this will cause a discontinuous regression loss and hinder the convergence of network training. To solve the issues, Circular Smooth Label \cite{yang2020arbitrary} and Densely Coded Labels \cite{yang2020dense} transform the angle regression task into a classification task to avoid the discontinuity of regression loss. SCRDet \cite{yang2019scrdet} adopts IoU-Smooth-L$_1$ loss to alleviate the sharp increase of loss caused by angle boundary. RSDet \cite{qian2019learning} utilizes a modulation loss function to solve the rotation-sensitive error of out-of-bounds angles.

There are also some methods adopt different oriented object representations to avoid the confusion issue of sequential label points under QBB representation \cite{liu2019omnidirectional,xu2020gliding,feng2020toso}.  Specifically, SBD \cite{liu2019omnidirectional} predicts key edges of the vertices, and uses an extra combination learning to determine the  quadrilateral. Gliding Vertex \cite{xu2020gliding} and TOSO \cite{feng2020toso} predict the horizontal bounding box of the object and the relative gliding offset to represent the quadrilateral bounding box.  

\begin{figure}[t]
\begin{center}
\includegraphics[width=1.0\linewidth]{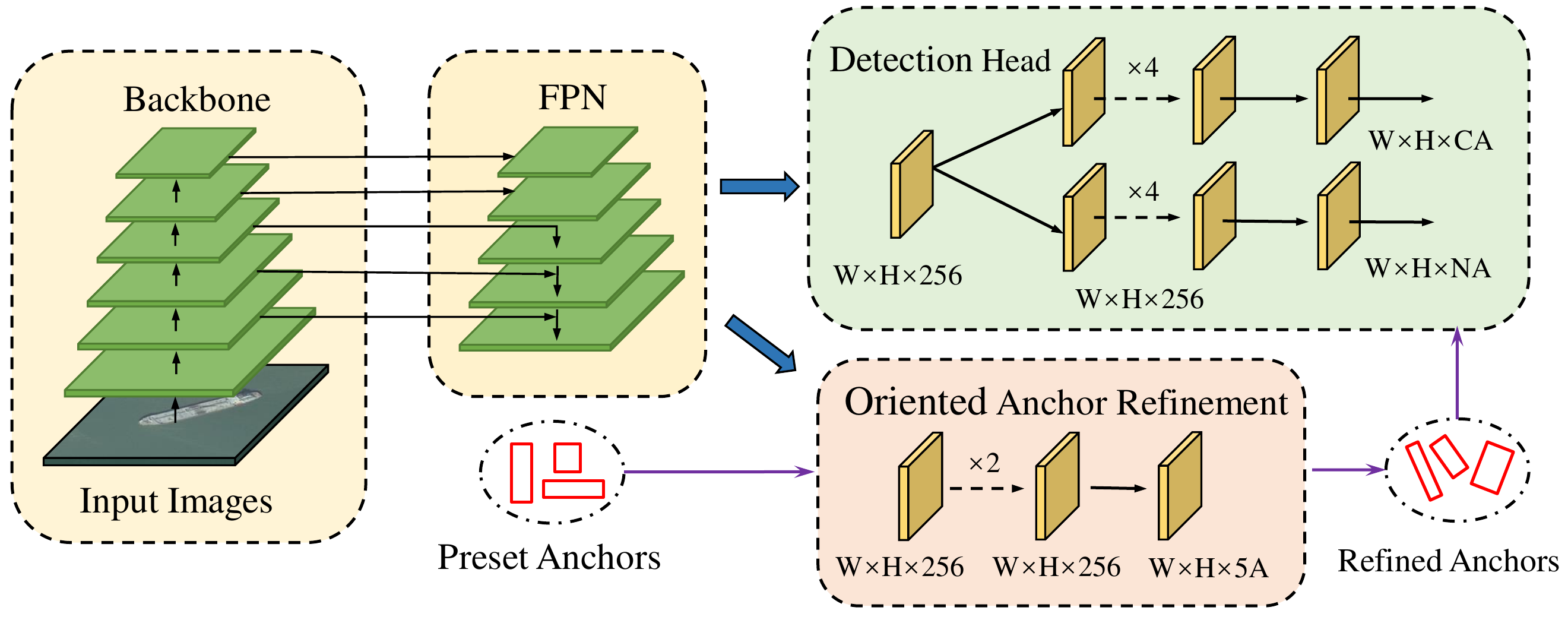}
\end{center}
   \caption{The overall structure of Cas-RetinaNet. $C$ represents the number of classes, and $A$ denotes the number of predefined anchors. $N$ is determined by the parameters of the representation, $N=5$ for OBB, while $N=8$ for QBB.}
\label{fig2}
\vspace{-8pt}
\end{figure}
However, these methods simply attemp to eliminate the problems caused by the nonunique definitions of oriented objects, but ignoring that these ambiguous representations are essentially equivalent local optima for regression optimization. Unlike the previous work, our method utilizes diverse representations to search for optimal regression strategy. In this case, the representation ambiguity not only no longer causes the various representation problems mentioned above, but also helps to achieve better performance.


\begin{figure}[t]
\begin{center}
\includegraphics[width=0.95\linewidth]{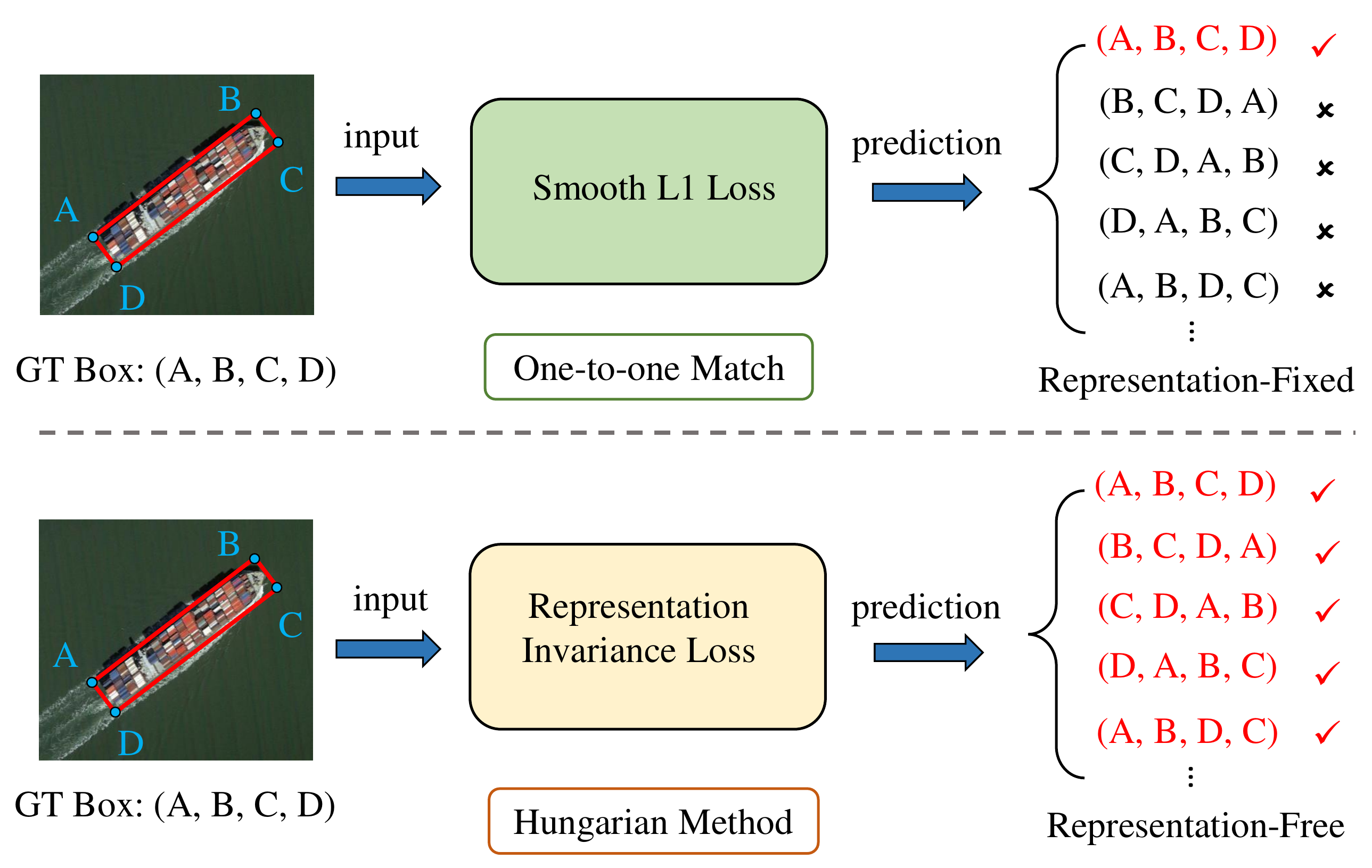}
\end{center}
   \caption{Comparison of regression results of RIL and smooth-L$_1$ loss under QBB. RIL treats ambiguity representations as equivalent local minima, allowing the detector to regress and output sequential-free predictions.}
\label{fig3}
\vspace{-8pt}
\end{figure}

\section{Representation Invariant Detector}
In this section, we first build a baseline and then design Representation Invariance Loss (RIL) for OBB methods and QBB methods, respectively. Then, we apply RIL to the baseline detector to construct Representation Invariant Detector (RIDet).

\subsection{Cas-RetinaNet for Orineted Object Detection}
We build the Cascaded RetinaNet (Cas-RetinaNet) as the baseline to achieve high-efficiency oriented object detection. Cas-RetinaNet uses horizontal preset anchors, and the Oriented Anchor Refinement Module (O-ARM) is adopted to obtain high-quality training samples. The overall structure of our model is shown in Fig.~\ref{fig2}. The Cas-RetinaNet is built based on RetinaNet \cite{lin2017focal}, in which FPN is applied to extract multi-scale features. The detection head performs classification and regression based on the high-quality anchors generated by O-ARM. 

The training loss of Cas-RetinaNet consists of three parts: classification loss, anchor refining loss, and bounding box regression loss. The loss function is as follows:
\begin{equation}
\small
\begin{split}
L = \frac{1}{N}&\sum_{i } FL\left(p_{i}, p_{i}^{*}\right) + \frac{1}{N_{p_1}}\sum_{i}  L_{ref}\left(\bm{t_{i}}, \bm{t_{i}^{*}}\right)  \\
&+ \frac{1}{N_{p_2}}\sum_{i}L_{RI}\left(\bm{b_{i}}, \bm{t_{i}^{*}}\right),
\end{split}
\end{equation} 
in which $FL$ denotes focal loss \cite{lin2017focal} for classification task, and $L_{ref}$ is smooth-L$_1$  loss for oriented anchor refinement. $p^*$ and $\bm{t^*}$ are ground-truth labels for classification, anchor refinement and box regression, while $p$, $\bm{t}$, $\bm{b}$ are corresponding predictions, respectively.  $N$ is the number of training samples, and $N_{p_1}$, $N_{p_2}$ denote the number of positives in the O-RAM stage and the detection stage, respectively. $L_{RI}$ is the representation invariance loss, which will be elaborated in the next sections.

\subsection{Quadrilateral Regression as Point Assignment}
Quadrilateral bounding boxes for most arbitrary-oriented objects are convex polygons, such as scene text \cite{yao2012detecting,karatzas2015icdar}, remote sensing objects \cite{xia2018dota,liu2017high,zhu2015orientation}, human faces \cite{shi2018real}, and retail products \cite{pan2020dynamic}. Therefore, the sequence of vertices is redundant to define a convex quadrilateral. Even so, the current detectors still learn the one-to-one matching between the given GT representation and the predictions as illustrated in the upper part of Fig.~\ref{fig3}.

\begin{figure}[t]
\begin{center}
\includegraphics[width=0.9\linewidth]{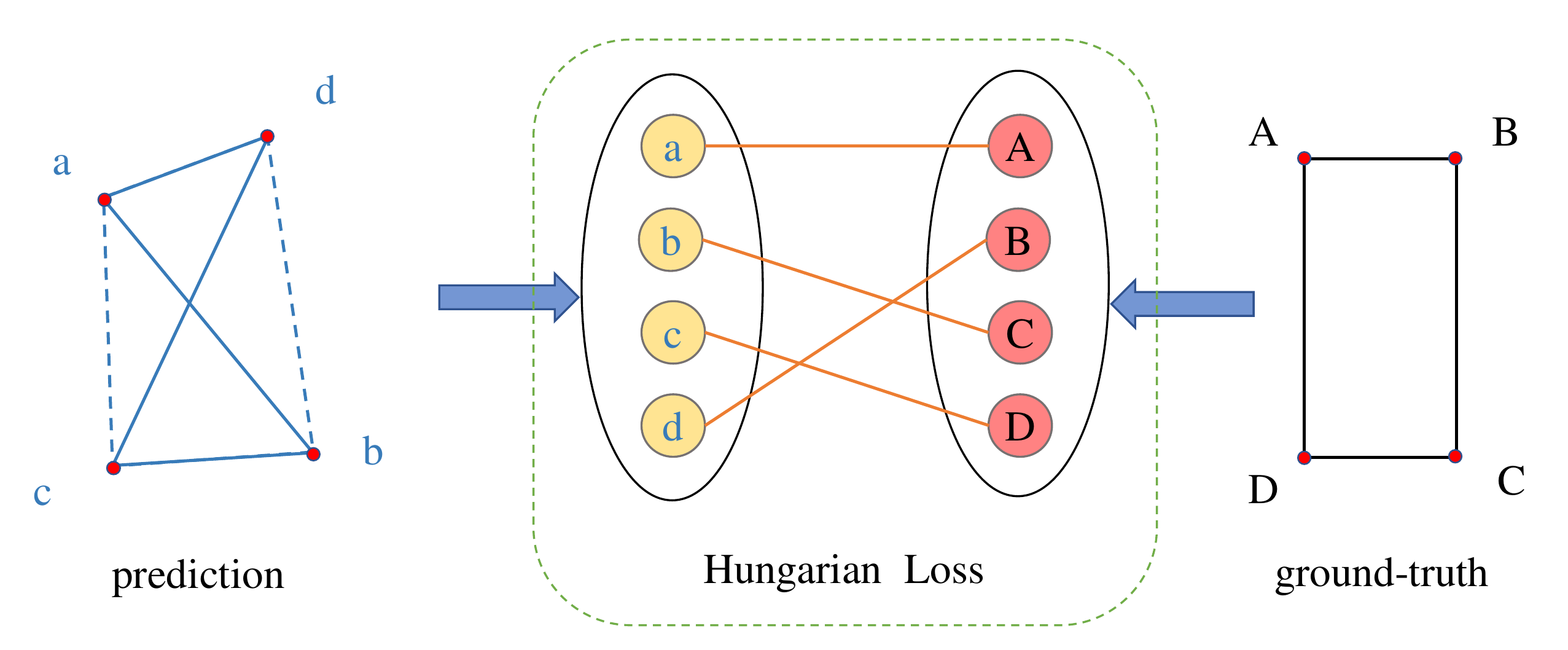}
\end{center}
   \caption{Illustration of matching between GT quadrilateral box and predicted quadrilateral box.}
\label{fig4}
\vspace{-8pt}
\end{figure}

Inspired by the Hungarian method for solving the assignment problem \cite{kuhn1955hungarian}, we treat the quadrilateral regression as a dynamic sequential-free points assignment problem. As shown in Fig.~\ref{fig4}, we denote the GT quadrilateral as an unordered  point set as $g= \{A,B,C,D\}$ and the predicted quadrilateral as $p=\{a,b,c,d\}$. The bounding box regression process can be transformed to the distance optimization between two point sets. We further use Hungarian loss to measure the distance between two sets, and thus the representation invariance loss (RIL) for QBB is as follows:
\begin{equation}  
L_{RI}(p,g) = \min _{\pi \in \Pi}\sum_{\substack{ v \in p, \ v^* \in g_{\pi}}} L_{reg}(v,v^*),
\end{equation}
in which $\Pi$ is permutations of $\{1,2,3,4\}$, it represents all feasible sequence of the quadrilateral. $\pi$ is a  permutation in $\Pi$, and $g_{\pi}$ represents a certain kind of quadrilateral.  $L_{reg}$ denotes smooth-L$_1$ loss here. 

With the RIL defined above, the network tends to search for the permutation with the minimal loss value among all ambiguous representations for bounding box optimization. For example, the vertices of the prediction in Fig.~\ref{fig4} are very close to the GT quadrilateral, but the unmatched point sequence leads to a large loss for the common rotation detectors. Our RIL can perform sequential-free auto-matching to achieve the optimal regression process and accurately predict the position of the vertices. Note that the training process is constantly changing, so the optimal regression strategy is not fixed, and it may switch between different sequences in $\Pi$ during training. RIL can adaptively select the most suitable representation according to the current optimization situation, and thus can achieve better convergence and performance.

The Hungarian method, which is representation invariant, no longer focuses on the sequence of points, and adaptively searches for the nearby local minimum in all possible permutations to achieve better convergence quality. Moreover, each four-point set determines at most one circumscribed convex quadrilateral. Therefore, in the inference stage, the network allows multiple predictions corresponding to ambiguous representations of the GT objects, as shown in Fig.~\ref{fig3}.

\begin{figure}[t]
	\subfigure[]{
		\includegraphics[width=0.22\textwidth]{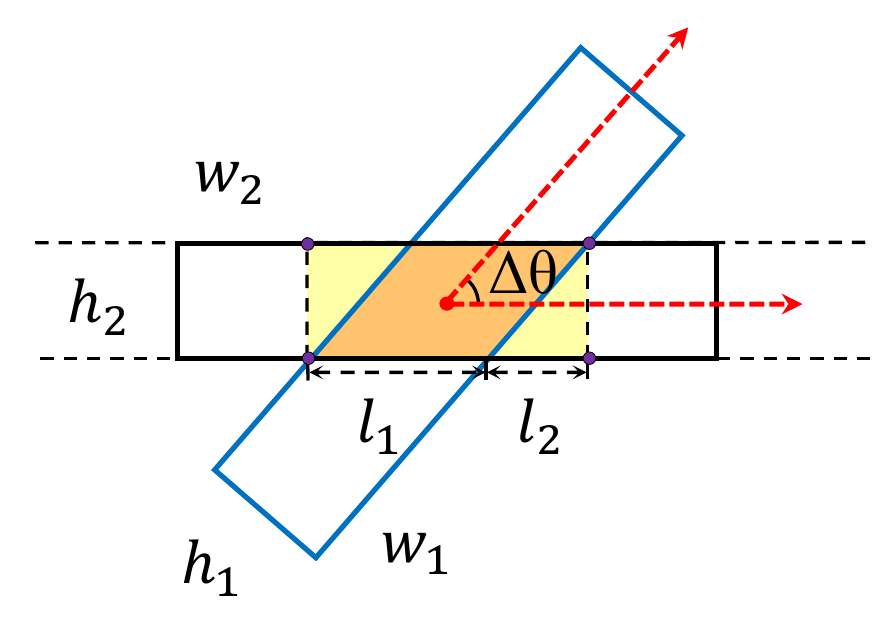}}\hspace{-1mm}
	\subfigure[]{
		\includegraphics[width=0.22\textwidth]{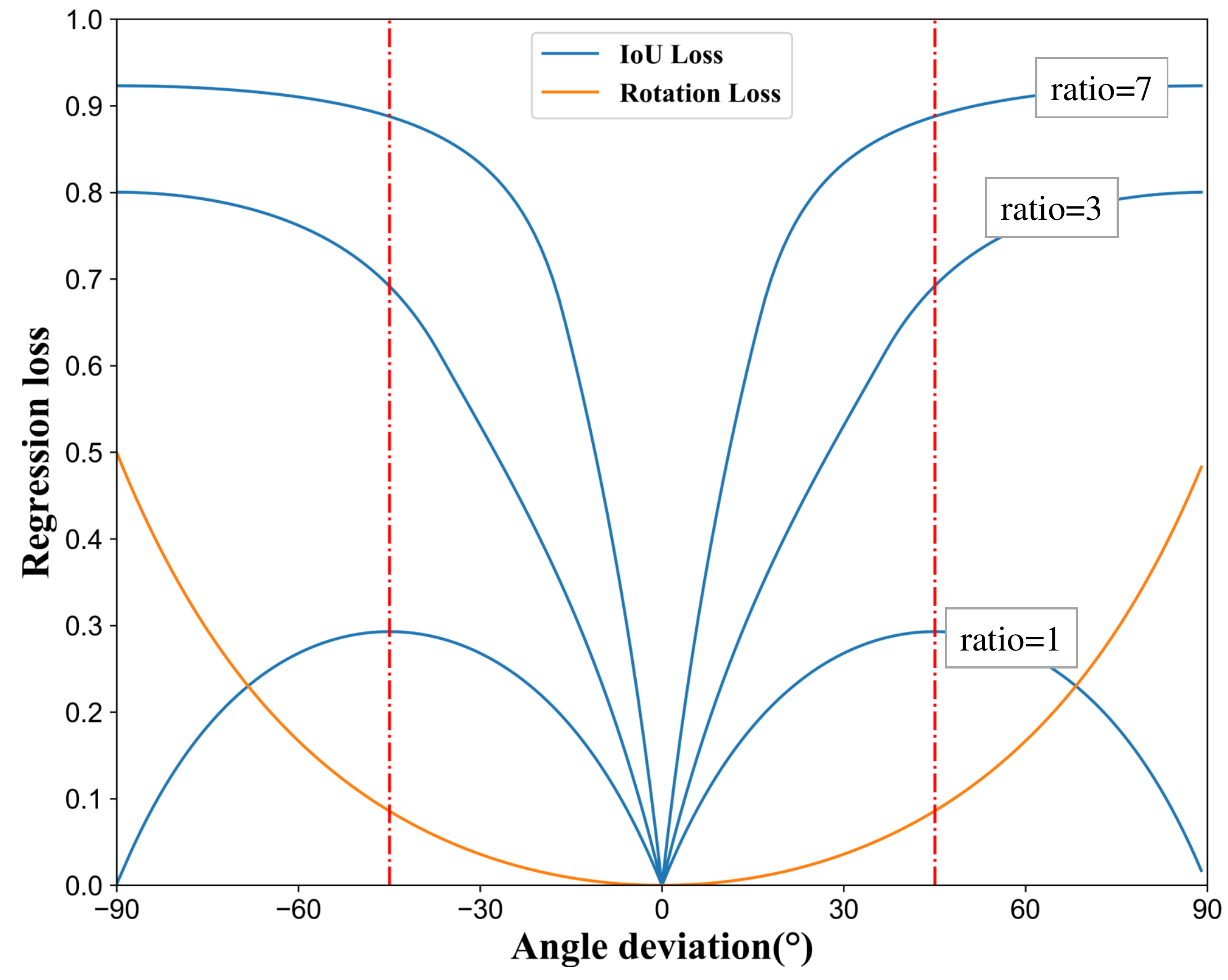}}
	\caption{Illustration of normalized angle mapping loss. (a) shows the definition of normalized angle mapping loss. (b) reveals its superiority for measuring angular deviation.}
	\label{fig5}
\end{figure}

\subsection{Out-of-bounds Matching for OBB} 

For the OBB, the boundary of the angle definition constrains the search space of regression loss, and thus the ambiguous representations cannot be effectively utilized to optimize the regression process. Based on the above observations, a Hungarian method based out-of-bounds matching strategy is proposed to solve the boundary constrains. Specifically, we transform the oriented bounding box regression into the adaptive optimal matching between the predictions and the representation space of GT.

Given an object $g$ denoted as $(cx, cy, w, h, \theta)$, its representation space $\Omega(g)$ is as follows:
\begin{equation} 
\Omega(g) = \{g_0,g_1,g_2,...,g_i\},
\end{equation}
in which the instances are defined as:
\begin{equation} 
\begin{aligned}
g_i\!=\!\left\{\!
\begin{array}{lcl}
(cx,cy,w, h,\theta+\frac{i\pi}{2}),    & {i\in \{2k|k \in \mathbb{Z}\}}\\
(cx,cy,h, w,\theta+\frac{i\pi}{2}),    & {i\in \{2k+1|k \in \mathbb{Z}\}}\\
\end{array} \right. 
\end{aligned}
\end{equation} 
Further the RIL for OBB is as follows:
\begin{equation} 
L_{RI}(p,g) =\min _{g_i \in \Omega(g)}L_{nrl}(p,g_i),
\end{equation}
where $p$ is the prediction, and $L_{nrl}$ denotes the normalized rotation loss for regression which will be elaborated next.  

RIL guides the prediction to match the optimal instance in $\Omega(g)$ to search for the nearby local optimum. Noted that 
 the periodicity of the angle leads to an infinite number of possible representations, it is impossible to exhaust all of them for the best matching. Therefore, we further propose a normalized  loss to reduce the dimensionality of $\Omega(g)$.

Given the anchor $(x_1, y_1, w_1, h_1, \theta_1)$ and GT box $(x_1, y_1, w_2, h_2, \theta_2)$ with overlapping center points as shown in Fig.~\ref{fig5}, the orientations of the two bounding boxes form an acute angle $\Delta \theta$. This angle can be approximated by the projection of its area and the direction of the reference axis. Taking the orientation of the GT as the reference line, $l_1$ and  $l_2$ are the projections of the overlapping parallelograms, and the intersection area $S_0$ and projected area $S_1$ are as follows:
\begin{equation} 
\begin{aligned}  
&l_1 = \frac{h_1}{\sin \theta}, \quad l_2 = \frac{h_2}{\tan \theta},\\
&S_0 = l_1 \cdot h_2, \quad  S_1 = (l_1 + l_2) \cdot h_2.
\end{aligned} 
\end{equation}
Then the IoU mapping of the angle is as:
\begin{equation} 
\begin{aligned} 
L_{\theta} &= \frac{S_0}{S_1}-0.5= \frac{1}{1 + \alpha \cdot \cos\theta}-0.5,
\end{aligned} 
\end{equation}
in which $\alpha =\text{min}(\frac{h_1}{h_2},\frac{h_2}{h_1})$, and $L_{\theta} \in [0, 0.5]$. Only when the heights of the anchor and the GT box are equal, and the angle deviation is equal to $0$, $L_{\theta}$ reaches the minimum value of $0$. With this metric, the representation space of $g$ can be simplified  into $\Omega(g) = \{g_0,g_1\}$. In addition, as shown in the right of Fig.~\ref{fig5}, $L_{\theta}$ is not sensitive to the aspect ratio of the bounding box itself, and it can converge steadily to objects of different shapes.

\begin{figure}[t]
	\subfigure[\label{fig6a}]{
		\includegraphics[width=0.23\textwidth]{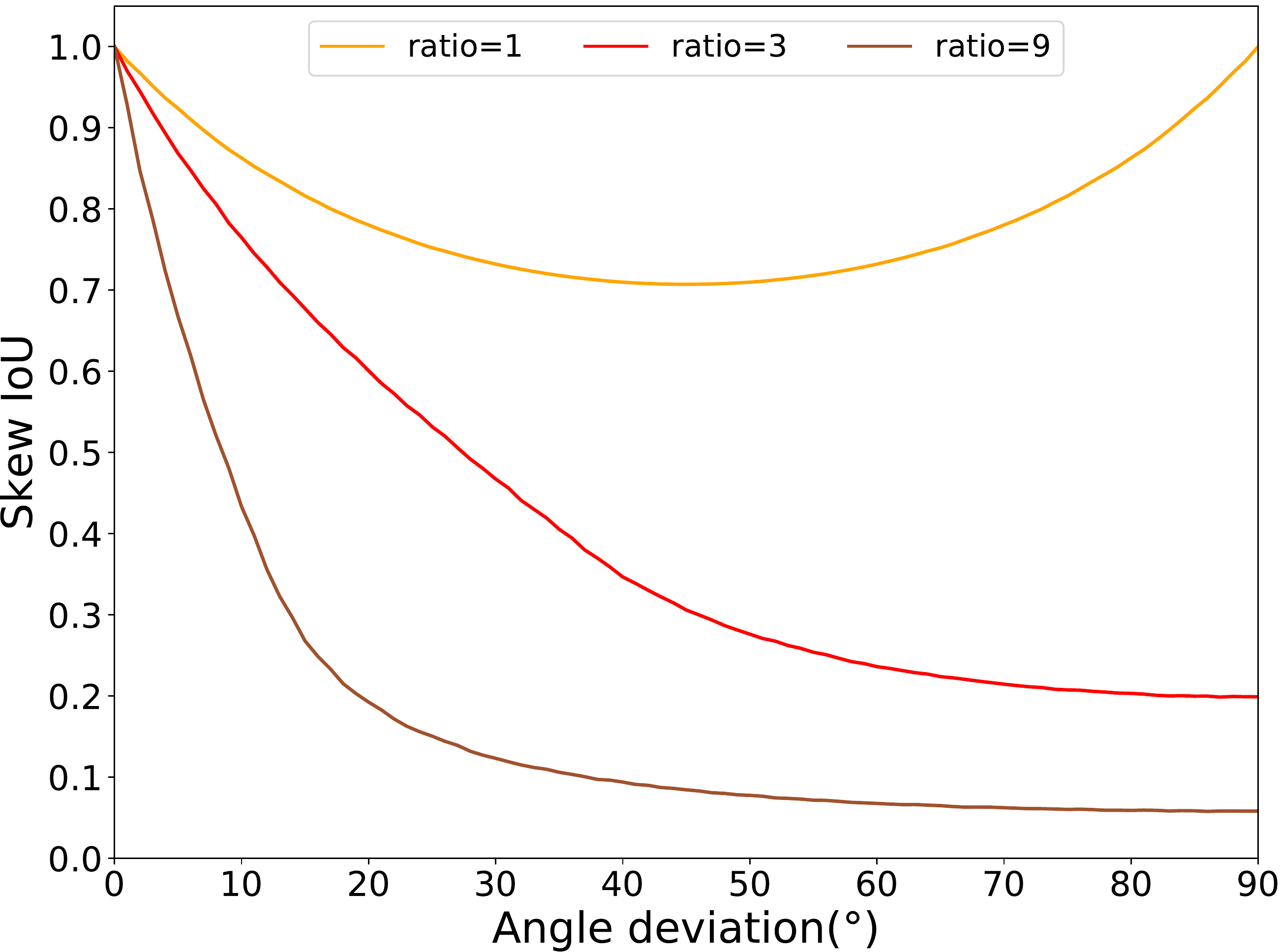}}\hspace{-1mm}
	\subfigure[\label{fig6b}]{
		\includegraphics[width=0.23\textwidth]{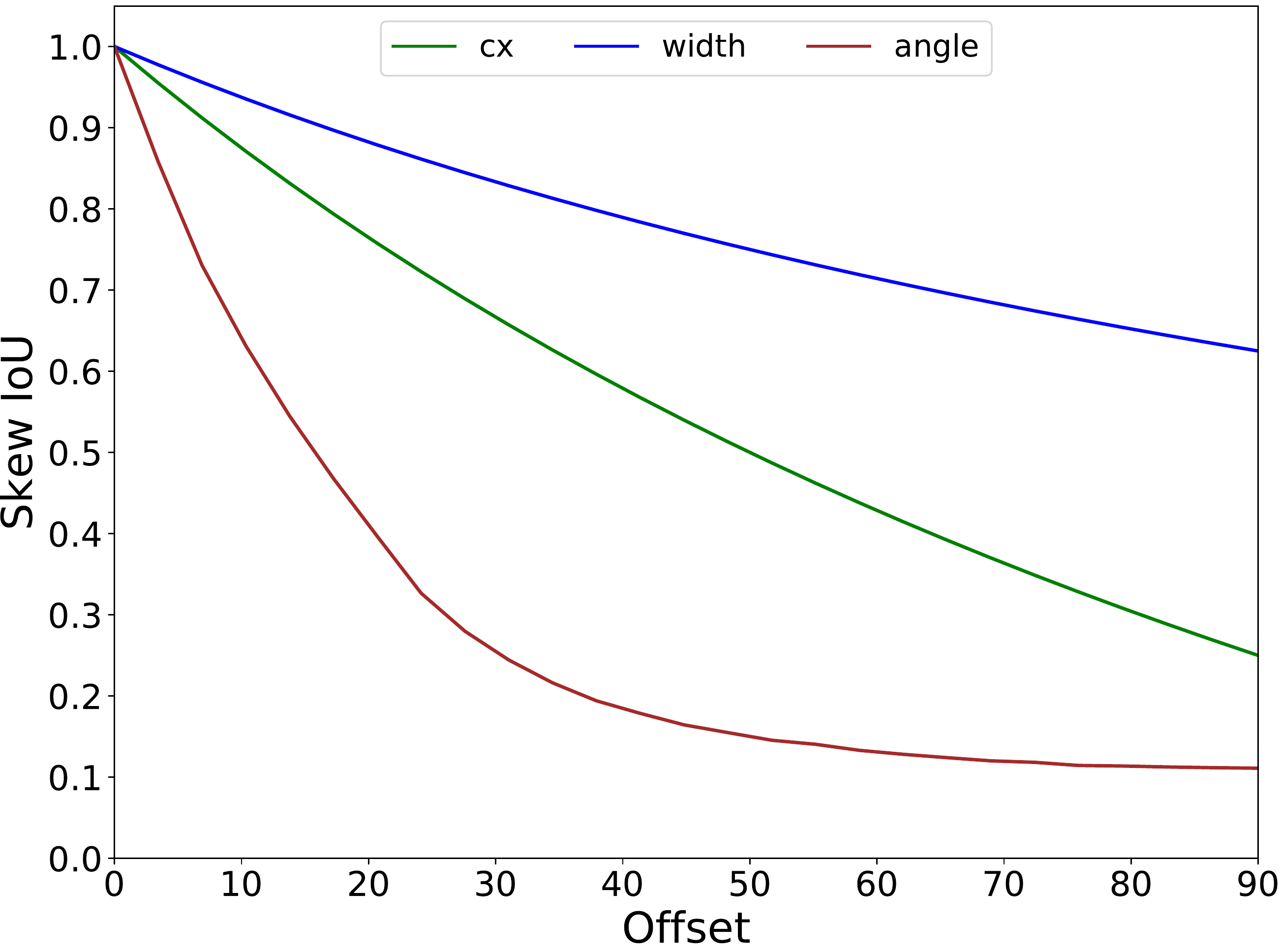}}
	\caption{Illustration of inconsistency between loss contribution and IoU variation. (a) shows the same offsets cause the inconsistent IoU changes due to the different aspect ratios of the GT box. (b) reveals that the same contribution of different variables to the loss cause different IoU changes.}
	\label{fig6}
	\vspace{-8pt}
\end{figure}

OBB representation requires five variables to define the oriented rectangular box. The deviation of IoU caused by the offset of a specic variable varies with the scale and aspect ratio of the GT box, as illustrated in the right of the Fig.~\ref{fig6a}. Besides, the IoU deviations caused by the same offset of different variables are also different, as shown in Fig.~\ref{fig6b}. As a result, there will be inconsistencies between the loss function and the detection metric, which has also been discussed in some previous work \cite{qian2019learning,yang2021rethinking}. Therefore, it is necessary to normalize the loss of variable offsets, just like $L_{\theta}$, to obtain a consistent and invariant loss metric.

We further constrain center coordinates and shape of the oriented bounding box to balance the loss contribution between different variables. The center loss $L_c$ is as follows:
\begin{equation} 
L_{c} =  \frac{ \left\| \bm{c_g} - \bm{c_a}\right\|_2}{ 0.5 \cdot \left\| \bm{d}\right\|_2}
\end{equation}
in which $\bm{c_g}$ and $\bm{c_a}$ are the center coordinates of GT and anchor respectively, $\bm{d}=(w,h)$ denotes the shape of the GT box. During the label assignment phase, only the anchors whose center points fall within the GT box may be selected as positive samples for regression \cite{zhang2020bridging}. In this way, $L_c$ is normalized in $[0,1]$. For example, small objects are more susceptible to inaccurate center point prediction with smooth-L$_1$ loss. Our normalized center loss is no longer sensitive to the shape of GT box, and thus it is conducive to the detection performance of small objects.

The same scale-sensitive problem also exists in shape regression. As discussed by Redmon \textit{et al.} \cite{redmon2016you}, small deviations in large boxes matter less than in small boxes. Therefore, we utilize the IoU of the horizontal box as the scale-independent shape loss, which is defined as follows:
\begin{equation} 
L_s =\min _{g_i \in \Omega(g)}hIoU(p,g_i),
\end{equation}
in which $hIoU$ calculates the IoU with two overlapping centers without angle consideration, it is defined as follows:
\begin{equation} 
hIoU(b_1, b_2) = \frac{\text{min}(w_1,w_2) \cdot \text{min}(h_1,h_2)}{\text{max}(w_1,w_2) \cdot \text{max}(h_1,h_2)}.
\end{equation}
Obviously, the range of the shape metric is also in $[0,1]$. The shape loss only evaluates the similarity of the shape and does not specify the width and height, which can speed up the convergence. The prediction of width and height is embedded in the angle loss to achieve an accurate description of the oriented rectangle box jointly.

Then the normalized rotation loss is as follows:
\begin{equation} 
L_{nrl} = L_{\theta} + L_c + L_s ,
\end{equation}
This novel RIL effectively uses entire representation space to optimize the regression process. Simultaneously, the normalized rotation loss can better alleviate the inconsistency of the contribution of different variables to the loss.

\section{Experiments}
\subsection{Datasets}
We conduct extensive experiments on multiple datasets to prove the effectiveness of the proposed method, including three remote sensing datasets: HRSC2016 \cite{liu2017high}, UCAS-AOD \cite{zhu2015orientation}, DOTA \cite{xia2018dota}, and two scene text datasets: ICDAR2015 \cite{karatzas2015icdar}, MSRA-TD500 \cite{yao2012detecting}.

HRSC2016 \cite{liu2017high} is a challenging ship detection dataset, which uses the oriented bounding box to annotate remote sensing ships. It contains 1,061 images with more than 20 categories of various ships. The dataset is divided into training, validation and test set, which contain 436, 181, 444 images respectively. UCAS-AOD \cite{zhu2015orientation} is an aerial image dataset for oriented aircraft and car detection. It contains 1,510 images, including 1,000 airplane images and 510 car images. Since no official dataset division strategy is provided, we randomly divide it into the training set, the validation set and the test set with a ratio of 5:2:3. DOTA \cite{xia2018dota} is the largest public dataset  with oriented bounding box annotations for object detection in remote sensing imagery. It contains totally 2,806 large size images (e.g. 4000$\times$4000 pixels). There are 15 categories in total.  Half of the images are the training set, 1/6 as the validation set, and 1/3 as the test set. Since that images in DOTA are too large, we crop images into 800$\times$800 patches with the stride set to 200.

ICDAR2015 \cite{karatzas2015icdar} is the scene text dataset for the incidental scene text challenge in ICDAR 2015 Robust Reading Competition. It contains 1,500 images, of which 1,000 for training and 500 for testing. MSRA-TD500 \cite{yao2012detecting} dataset consists of 300 training images and 200 testing images.  It contains text in both English and Chinese. The text regions are annotated with oriented bounding boxes.

\begin{table}[t]
	\begin{center}
	\footnotesize
	\renewcommand\arraystretch{1.1}
	\setlength{\tabcolsep}{2.5mm}{
	\begin{tabular}{{c|ccc|ll}}
		\toprule
		Backbone& OM    &   NL    &  BC     & IC15 & M500  \\ 
		\hline
		ResNet-50 & & & &										 75.9	&		  76.1 \\ 
		ResNet-50 & \checkmark & & &							 \multicolumn{1}{>{\columncolor{gray!10}}l}{76.5 \textbf{\textcolor[rgb]{0,0.6,0}{\scriptsize (+0.6)}}}	&		  \multicolumn{1}{>{\columncolor{gray!10}}l}{76.6 \textbf{\textcolor[rgb]{0,0.6,0}{\scriptsize (+0.5)}}}\\
		ResNet-50 & \checkmark &\checkmark &&					 \multicolumn{1}{>{\columncolor{gray!20}}l}{77.2 \textbf{\textcolor[rgb]{0,0.6,0}{\scriptsize (+1.3)}}}	&		  \multicolumn{1}{>{\columncolor{gray!20}}l}{77.5 \textbf{\textcolor[rgb]{0,0.6,0}{\scriptsize (+1.4)}}} \\
		ResNet-50 & \checkmark &\checkmark &\checkmark & \multicolumn{1}{>{\columncolor{gray!30}}l}{ \textbf{77.6 \textcolor[rgb]{0,0.6,0}{\scriptsize (+1.7)}}}	&\multicolumn{1}{>{\columncolor{gray!30}}l}{ \textbf{78.2 \textcolor[rgb]{0,0.6,0}{\scriptsize (+2.1)}}}\\
		\bottomrule
	\end{tabular}
		}
		\end{center}
\caption{Evaluation of different components of RIL for OBB. OM, NL, and BC represent optimal matching, normalized loss, and bounded center, respectively. IC15 denotes ICDAR2015 dataset, and M500 is MSRA-TD500 dataset.}
	\label{tab1}
\end{table}

\subsection{Implementation Deteails}

We use the Cas-RetinaNet as the baseline model, and RIL is applied to the baseline to build a Representation Invariant Detector (RIDet). It is further divided into RIDet-O for OBB and RIDet-Q for QBB. We use Pytorch \cite{paszke2019pytorch} to implement the above methods.

All images are resized to 416$\times$416 or 800$\times$800 for training and testing. Horizontal anchors are preset with aspect ratios of $\{0.5, 1, 2\}$, and scales of $\{2^0,2^{1/3},2^{2/3}\}$. The IoU thresholds used to select training samples are 0.4 in O-RAM and 0.5 in the detection stage. We use Adam optimizer for training with the initial learning rate set to 1e-4. We train the models in 100 epochs for HRSC2016, 24 epochs for DOTA, and 40 epochs for UCAS-AOD, ICDAR2015 and MSRA-TD500. All models are trained on RTX 2080 Ti GPUs with batch size set to 8. Random flip, rotation, and HSV colour space transformation are adopted for data augmentation. 

\subsection{Ablation Study}

\subsubsection{Evaluation of normalized rotation loss for OBB}
Experiments conducted on the two scene texts datasets confirmed the effectiveness of RIL under OBB representation. The experimental results are shown in Tab. \ref{tab1}. The baseline model is the Cas-RetinaNet detailed above.  The regression head outputs the oriented rectangular bounding box in the experiments in this section. Note that we do not use complex data augmentation in this part.

RIDet-O achieves the F-measure of 77.6\% and 78.2\% on the ICDAR2015 and MSRA-TD500, which are improved by 1.7\% and 2.1\% compared to the baseline model, respectively. The optimal matching strategy allows the out-of-bounds predictions to search for the optimal local minima for regression, which is conducive to network convergence, and thus better performance can be achieved. It improves the F-measure by 0.6\% and 0.5\%. The normalized rotation loss further anchieves the increase of 0.7\% and 0.9\% on two scene text datasets, which balances the contribution of different variables and alleviates the inconsistency between loss and detection metric. Finally, selecting the anchors with the center point inside the GT box for training improves by 0.4\% and 0.7\%. These high-quality positives not only provide better priori about the position of the center points but also help constrain the range of center loss, which has also been demonstrated in ATSS \cite{zhang2020bridging}.

\begin{table}[t]
	\footnotesize
	\renewcommand\arraystretch{1.}
	\setlength{\tabcolsep}{1.8mm}{
	\begin{center}
		\begin{tabular}{l|ccc|c}
			\toprule
			Method   & Image Res.     & BBox       & RIL  & mAP     \\ 
			\hline
			\multirow{2}{*}{RetinaNet \cite{lin2017focal}} & \multirow{2}{*}{512$\times$512}& \multirow{2}{*}{OBB}& $\times$  &82.2\\ 
			&&&\checkmark& \multicolumn{1}{>{\columncolor{gray!20}}l}{ \textbf{84.5 \textcolor[rgb]{0,0.6,0}{\scriptsize (+2.3)}}} \\ 
			\hline
			\multirow{2}{*}{Cas-RetinaNet \cite{lin2017focal}} &\multirow{2}{*}{416$\times$416} &\multirow{2}{*}{QBB}    &$\times$  &84.3\\ 
			&&&\checkmark & \multicolumn{1}{>{\columncolor{gray!20}}l}{ \textbf{86.3 \textcolor[rgb]{0,0.6,0}{\scriptsize (+2.0)}}} \\ 
			\hline
			\multirow{2}{*}{CFC-Net \cite{ming2021cfc}}  &\multirow{2}{*}{416$\times$416}   &\multirow{2}{*}{OBB} &$\times$   & 85.2\\ 
			&&&\checkmark  & \multicolumn{1}{>{\columncolor{gray!20}}l}{ \textbf{87.3 \textcolor[rgb]{0,0.6,0}{\scriptsize (+1.9)}}}  \\ 
			\hline
			\multirow{2}{*}{DAL \cite{ming2020dynamic}}  & \multirow{2}{*}{384$\times$384}  &\multirow{2}{*}{OBB}  &$\times$   & 87.1\\ 
			&&&\checkmark  & \multicolumn{1}{>{\columncolor{gray!20}}l}{ \textbf{88.5 \textcolor[rgb]{0,0.6,0}{\scriptsize (+1.4)}}}  \\ 
			\hline
			\multirow{2}{*}{S$^2$A-Net \cite{han2021align}}  & \multirow{2}{*}{416$\times$416}  &\multirow{2}{*}{OBB}  &$\times$   & 88.3\\ 
			&&&\checkmark  & \multicolumn{1}{>{\columncolor{gray!20}}l}{ \textbf{89.1 \textcolor[rgb]{0,0.6,0}{\scriptsize (+0.8)}}} \\ 
			\bottomrule
	\end{tabular}
	\end{center}
	}
\caption{Performance improvement of RIL on different methods on HRSC2016 dataset.}
\label{tab2}
\end{table}

\subsubsection{Evaluation on different models}
We further conduct experiments on different detectors to verify the generalization of RIL. The experimental results on HRSC2016 \cite{liu2017high} and DOTA \cite{xia2018dota} are shown in Tab. \ref{tab2} and Tab. \ref{tab5}, respectively. 

Experimental results in Tab. \ref{tab2} show that RIL improves the mAP by 2.3\% on RetinaNet with OBB, and 2.0\% on Cas-RetinaNet with QBB on HRSC2016 . It proves that the strategy of transforming the bounding box regression task into the optimal matching process between predction and the representation space of GT can be beneficial in different representation methods. The performance in Tab. \ref{tab5} further confirms this point. RIDet-Q and RIDet-O achieved 2.26\% and 1.81\% improvement on the corresponding baseline detector, respectively. S$^2$A-Net \cite{han2021align} is an advanced rotation detector that achieves 76.38\% mAP on the DOTA . Integrating RIL to S$^2$A-Net improves performance by 1.24\% to achieve the mAP of 77.62\%, which further proves the generalization and superiority of our method.
 

DAL \cite{ming2020dynamic} and CFC-Net \cite{ming2021cfc} are also currently advanced detectors that achieve accurate oriented object detection. But these detectors will still be confused by ambiguous representations of GT. After training with RIL, the detection performance is further improved by 1.9\% and 1.4\% on DAL \cite{ming2020dynamic} and CFC-Net \cite{ming2021cfc}, respectively. It indicates that as long as there is representation ambiguity, our method can obtain consistent performance gains.

Extensive experiments on different models and representations prove that our method can effectively optimize the representation learning method for the rotation detectors and achieve stable performance improvements.

\begin{table}[t]
\footnotesize
\renewcommand\arraystretch{1.}
\begin{center}
\setlength{\tabcolsep}{1.5mm}{
	\begin{tabular}{c|cc|c|cc}
		\toprule
		BBox  & Models   & mAP &  BBox  & Model  & mAP   \\
		\hline
		\multirow{4}{*}{OBB} &  IoU-Smooth  \cite{yang2019scrdet}  &82.59
			& \multirow{4}{*}{QBB} 	&  MR Loss  \cite{qian2019learning} & 86.50 \\ 
		&  MR Loss \cite{qian2019learning} 				    & 83.60  &										
			&  Gliding Vertex \cite{xu2020gliding}   &	88.20\\
		&  DCL  	\cite{yang2020dense}					    & 89.46  &								  	   		&  BBAVectors  	\cite{yi2020oriented}	&	88.60\\
		&  RIDet-O  	& \textbf{89.63}  &		&  RIDet-Q  	&	\textbf{89.10}\\
\bottomrule
	\end{tabular}}
\end{center}
\caption{Comparison with some related methods on HRSC2016.}
\label{tab3}
\end{table}

\subsubsection{Comparison with related methods}
We have conducted comparisons of the performance of different models in related work on the HRSC2016 dataset , and the results are shown in Tab. \ref{tab3}. 

RIDet-O achieves the highest mAP of 89.63\%  among the compared methods. IoU-Smooth \cite{yang2019scrdet} and MR Loss \cite{qian2019learning} solve the loss discontinuity caused by multiple representations and achieve the mAP of 82.59\% and 83.60\%, respectively. However, the weight of skew IoU in IoU-Smooth Loss \cite{yang2019scrdet} is not differentiable, and MR Loss  \cite{qian2019learning}  only considers limited redundant representations.
DCL \cite{yang2020dense}	 converts the angle regression into  fine-grained angle classification  to eliminate the problems caused by the angle boundary, and thus achieving the mAP of 89.46\%.  But it brings additional computational overhead and cannot be directly used to other detectors. Compared with these works with OBB, our method can be directly applied to mainstream rotation detectors to achieve the optimal regression and balanced training without any inference overhead.

Among the compared quadrilateral detectors, RIDet-Q achieves the highest mAP of 89.10\%, which proves the effectiveness of our RIL. MR Loss \cite{qian2019learning} achieves the mAP of 86.5\%, which also consider the order of the vertices for QBB. The main difference is that MR loss traverses the four ordered point sequences to find the suitable representation, which make it limited to the ordered quadrilateral representation, and the traversal of the possible permutations is both time-consuming and insufficient. The offset prediction in Gliding Vertex \cite{xu2020gliding} depends on the accurate prediction of the corresponding horizontal box, which further leads to  optimization difficulties. In contrast, RIDet-Q is a completely sequence-free quadrilateral detector. It solves the problem of representation ambiguity via dynamically searching for optimal equivalent representations, and thus improves the detection performance.

\begin{table}[t]
	\footnotesize
	\renewcommand\arraystretch{1.}
	\begin{center}
	\begin{tabular}{c|cc|c}
		\toprule
		Methods & Backbone & Size  & mAP \\
		\hline
		\multicolumn{1}{l}{\textit{Two-stage:}}&\multicolumn{3}{l}{}  \\
		\hline		
		$\text{R}^2\text{CNN}$ \cite{jiang2017r2cnn}        & ResNet101   &800$\times$800      & 73.07 \\
		RRPN \cite{ma2018arbitrary}        &  ResNet101  & 800$\times$800      & 79.08  \\  
		\text{R}$^2$\text{PN} \cite{zhang2018toward}        &  VGG16      &  ---                 &  79.60 \\ 
		RoI Trans.  \cite{ding2019learning}&  ResNet101  & 512$\times$800     &   86.20\\ 
		Gliding Vertex \cite{xu2020gliding}&  ResNet101  & 512$\times$800      &  88.20 \\ 
		OPLD \cite{song2020learning}  &  ResNet50  & 1024$\times$1333      &  88.44 \\
		DCL \cite{yang2020dense}  &  ResNet101  & 800$\times$800      &  89.46 \\
		\hline
		\multicolumn{1}{l}{\textit{Single-stage:}}&\multicolumn{3}{l}{}  \\
		\hline
		RetinaNet \cite{lin2017focal} & ResNet50 & 416$\times$416  & 80.81 \\ 
		RRD \cite{liao2018rotation}   &  VGG16   & 384$\times$384  & 84.30 \\ 
		RSDet \cite{qian2019learning} & ResNet50 & 800$\times$800  & 86.50 \\ 
		BBAVector \cite{yi2020oriented}&  ResNet101& 608$\times$608 & 88.60\\
		$\text{R}^3\text{Det}$ \cite{yang2019r3det}   &  ResNet101  & 800$\times$800     &  89.26 \\ 

		RIDet-O (Ours)     &  ResNet18   & 416$\times$416      & 89.25\\
		RIDet-O (Ours)     &  ResNet50   & 416$\times$416      & 89.47	\\
		RIDet-Q (Ours)    &  ResNet101  & 800$\times$800        &  89.10 \\
		RIDet-O (Ours)    &  ResNet101  & 800$\times$800        & \textbf{89.63}  \\
		\bottomrule
	\end{tabular}
	\end{center}
\caption{Comparisons with advance methods on HRSC2016.}	
	\label{tab4}
\end{table}

\begin{table*}[t]
	\footnotesize
	\renewcommand\arraystretch{1.}
	\begin{center}
	\setlength{\tabcolsep}{1.2mm}{
	\begin{tabular}{c|c|ccccccccccccccc|c}
		\toprule
		Methods & Backbone  & PL & BD & BR & GTF & SV & LV & SH & TC & BC & ST & SBF & RA & HA & SP & HC & mAP \\
		\hline
		 \multicolumn{1}{l}{\textit{Two-stage:}} &\multicolumn{17}{l}{}  \\
		\hline		
		RRPN \cite{ma2018arbitrary}	   & R-101                   & 88.52  &71.20  &31.66  &59.30   &51.85  &56.19  &57.25  &90.81  &72.84  &67.38  &56.69  &52.84  &53.08  &51.94  &53.58  &61.01\\ 
		ICN \cite{azimi2018towards} 	   &  R-101                  & 81.36  &74.30  &47.70  &70.32   &64.89  &67.82  &69.98  &90.76  &79.06  &78.20  &53.64  &62.90  &67.02  &64.17  &50.23  &68.16\\
		RoI Trans.  \cite{ding2019learning}& R-101                 & 88.64  &78.52  &43.44  &75.92   &68.81  &73.68  &83.59  &90.74  &77.27  &81.46  &58.39  &53.54  &62.83  &58.93  &47.67  &69.56\\
		CAD-Net \cite{zhang2019cad}      &  R-101                 & 87.80  &82.40  &49.40  &73.50   &71.10  &63.50  &76.70  &$\textbf{90.90}$  &79.20  &73.30  &48.40  &60.90  &62.00  &67.00  &62.20  &69.90 \\ 
		O$^2$-DNet \cite{wei2020oriented} &  H-104  & 89.31  &82.14  &47.33  &61.21   &71.32  &74.03  &78.62  &90.76  &82.23  &81.36  &60.93  &60.17  &58.21  &66.98  &61.03  &71.04\\

		SCRDet \cite{yang2019scrdet}	 &  R-101                  & $\textbf{89.98}$  &80.65  &52.09  &68.36   &68.36  &60.32  &72.41  &90.85  &$\textbf{87.94}$  &86.86  &65.02  &66.68  &66.25  &68.24  &65.21  &72.61\\ 


		\hline
		\multicolumn{1}{l}{\textit{Single-stage:}}&\multicolumn{17}{l}{}  \\
		\hline
		DRN \cite{pan2020dynamic}        &  H-104                 & 88.91  &80.22  &43.52  &63.35   &73.48  &70.69  &84.94  &90.14  &83.85  &84.11  &50.12  &58.41  &67.62  &68.60  &52.50  &70.70  \\ 
		DAL \cite{ming2020dynamic} &  R-101     &88.61     &79.69     &46.27     &70.37     &65.89     &76.10     &78.53     &90.84     &79.98     &78.41     &58.71     &62.02     &69.23     &71.32     &60.65     &71.78\\	
		RSDet  \cite{qian2019learning} &  R-101  & 89.80   &$\textbf{82.90}$   &48.60   &65.20   &69.50   &70.10   &70.20   &90.50   &85.60   &83.40   &62.50   &63.90   &65.6 0  &67.20   &$\textbf{68.00}$   &72.20\\
		BBAVector \cite{yi2020oriented} &  R-101 &88.35  &79.96  &50.69  &62.18  &78.43  &78.98  &87.94  &90.85  &83.58   &84.35   &54.13   &60.24   &65.22   &64.28   &55.70   &72.32 \\
		CFC-Net \cite{ming2021cfc} &  R-50        &89.08     &80.41     &52.41     &70.02     &76.28     &78.11     &87.21     &90.89     &84.47     &85.64     &60.51     &61.52     &67.82     &68.02     &50.09     &73.50\\
		R$^3$Det \cite{yang2019r3det}   &  R-152        & 89.49  &81.17  &50.53  &66.10   &70.92  &78.66  &78.21  &90.81  &85.26  &84.23  &61.81  &63.77  &68.16  &69.83  &67.17  &73.74 \\ 
		S$^2$A-Net \cite{han2021align}      &  R-50         &89.07  &82.22   &53.63  &69.88   &$\textbf{80.94}$  &$\textbf{82.12}$  &88.72  &90.73  &83.77  &86.92  &63.78  &$\textbf{67.86}$  &76.51  &73.03  &56.60  &76.38\\ 
		\hline		
		Baseline-Q                    &  R-101       &88.65  &75.53 & 43.68 	 &66.48 	 &62.92 	 &78.59  &	79.42 	 &90.84  &79.97 	 &74.86 	 &62.95  &	58.77  &	68.29  &	72.06 	 &54.83  &70.52\\

		RIDet-Q                    &  R-101       &87.38  &75.64 & 44.75 	 &70.32 	 &77.87 	 &79.43  &	87.43 	 &90.72  &81.16 	 &82.52 	 &59.36  &	63.63  &	68.11  &	71.94 	 &51.42  &72.78\\ 

		Baseline-O                    &  R-101         & 88.36	&75.31	&46.33	&66.59 	&76.88 	&77.88	&87.23	&90.51	&82.73	&83.84	&60.89	&60.73 	&67.29   &72.01&56.91 &72.89\\

        RIDet-O                    &  R-101         & 88.94	&78.45	&46.87	&72.63 	&77.63 	&80.68	&88.18	&90.55	&81.33	&83.61	&$\textbf{64.85}$	&63.72 	&73.09   &73.13&56.87 &74.70\\
		S$^2$A-Net + RIL                             &  R-50         &89.31   &80.77  	 &$\textbf{54.07}$  &$\textbf{76.38}$   &79.81  &81.99 &$\textbf{89.13}$  &90.72  &83.58   &$\textbf{87.22}$  &64.42  &67.56  &$\textbf{78.08}$  &$\textbf{79.17}$  &62.07  &$\textbf{77.62}$  \\ 
		\bottomrule
	\end{tabular}}
		\end{center}
\caption{ Performance evaluation of OBB task on DOTA dataset. H-104 is Hourglass-104. R-50 and R-101 denote ResNet-50 and ResNet-101, respectively.}
\label{tab5}
\end{table*}

\subsection{Comparison with the State-of-the-Art}

\subsubsection{Results on HRSC2016}
The HRSC2016 dataset \cite{liu2017high} contains lots of oriented  ships. As shown in Tab. \ref{tab4}, RIDet outperforms other single-stage detectors and even some recent advanced two-stage detectors such as DCL \cite{yang2020dense}, RoI Transformer\cite{ding2019learning}. RIDet-O and RIDet-O achieve the mAP of 89.10\%, and 89.63\%, respectively. Especially when the images are resized to 416$\times$416 with lightweight ResNet-18 as the backbone, RIDet-O still achieve competitive performance of 89.25\%.  

\begin{figure}[t]
	\subfigure[\label{fig7a}]{
		\includegraphics[width=0.23\textwidth]{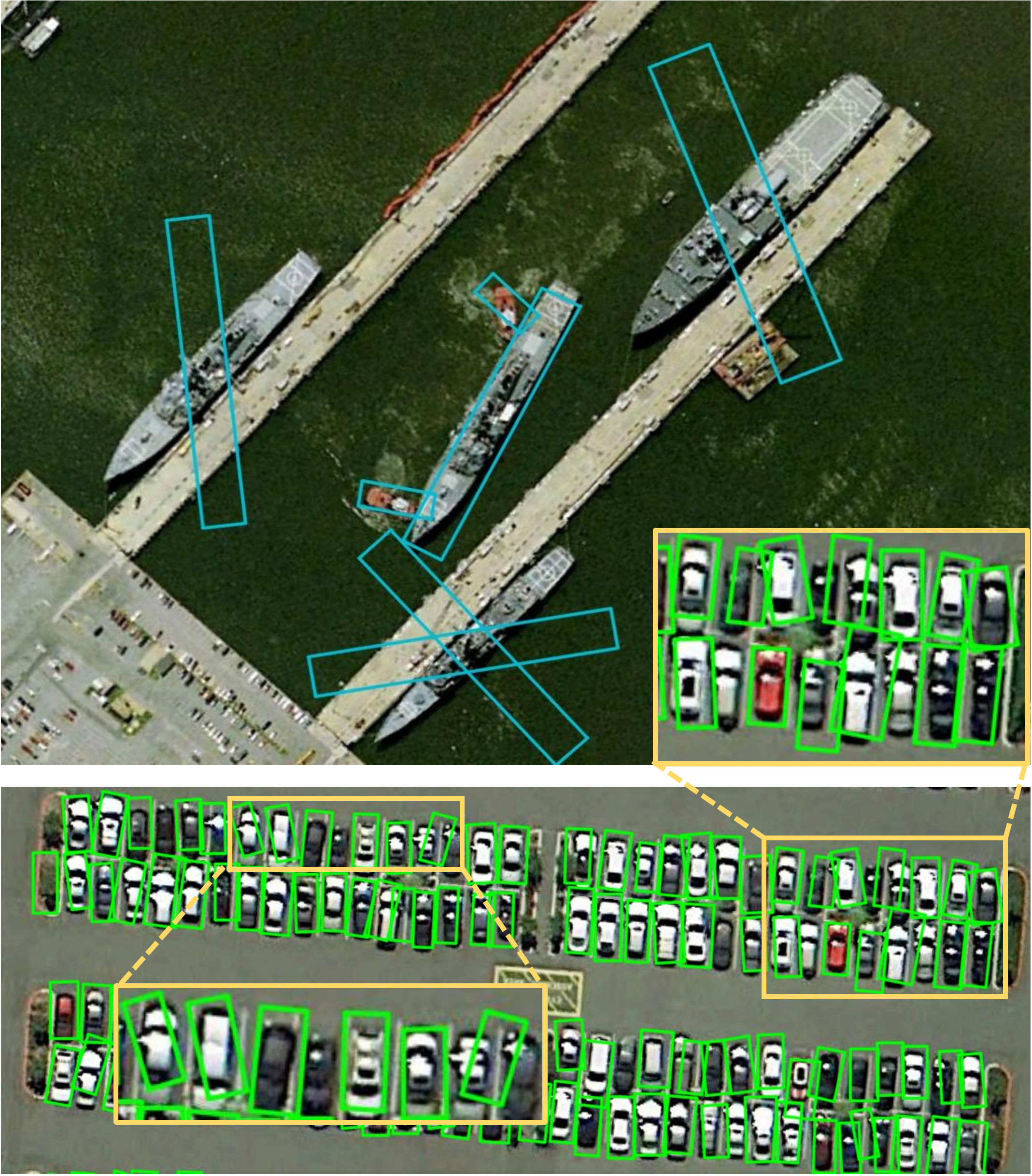}}\hspace{-1mm}
	\subfigure[\label{fig7b}]{
		\includegraphics[width=0.23\textwidth]{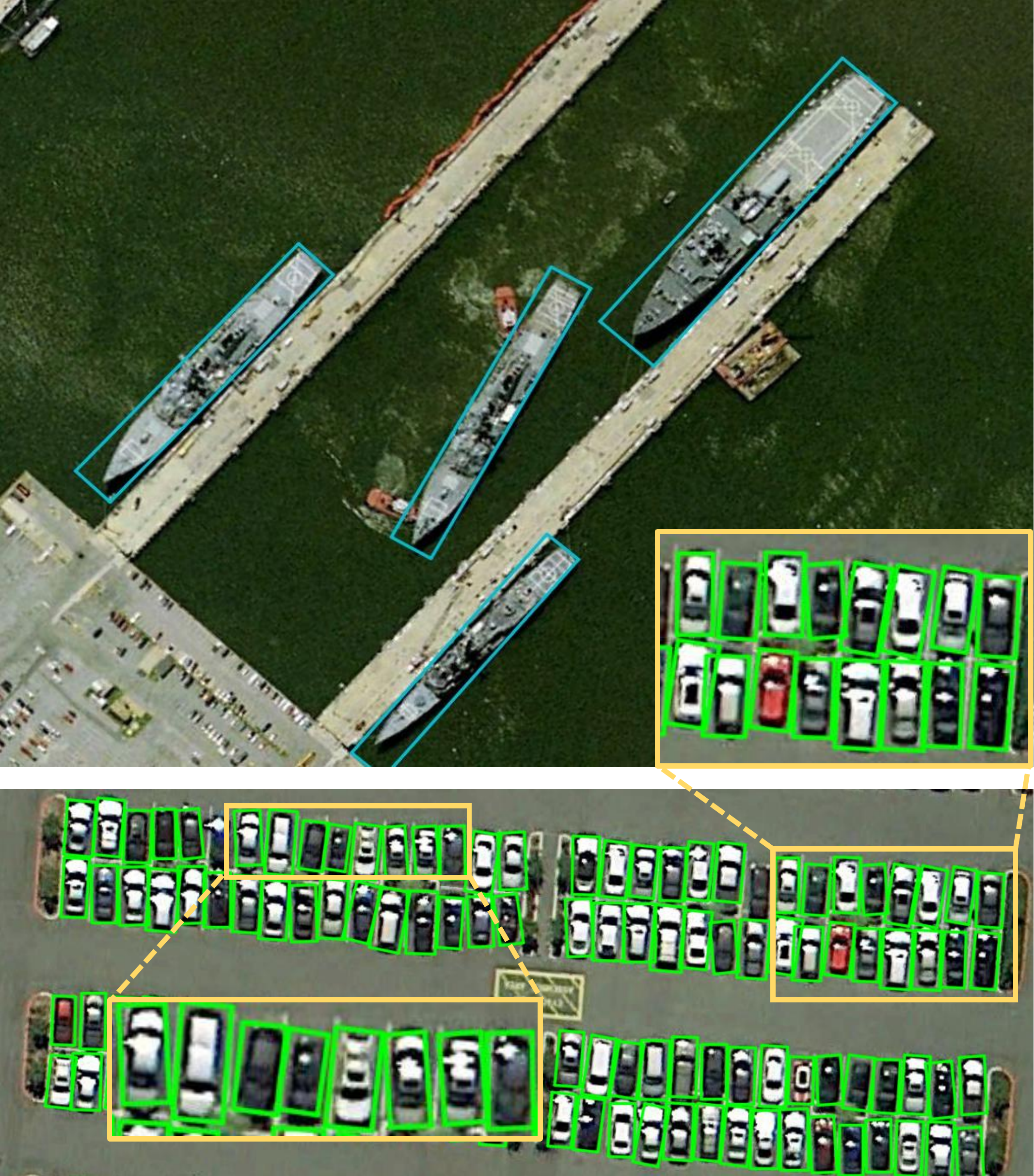}}
	\caption{Comparison of detection results between detectors with smoth-L$_1$ loss (a) and RIL (b) .}
\label{fig7}
\end{figure}

The visualization results in Fig.~\ref{fig7a} show that the convergence of the angle regression is slow in the baseline model which only learns a specific representation. RIDet treats all feasible representations as equivalent local optimums. Therefore the regression loss converges faster, and better performance can be achieved under the same number of iterations, as shown in Fig.~\ref{fig7b}.

\subsubsection{Results on DOTA}
We compared the proposed method with other  state-of-the-art detectors on the DOTA dataset.  The ground-truth labels of the  test set are not publicly available, and thus the results in Tab. \ref{tab5} are obtained by submitting the detection results to the official DOTA evaluation server. 

As shown in Tab. \ref{tab5}, RIDet-Q and RIDet-O achieve mAP of 72.78\% and 74.70\%, respectively. Note that the baseline model we used is the naive one-stage detector Cas-RetinaNet. After using RIL, the performance of RIDet can be comparable to many state-of-the-art methods,  such as R$^3$Det  \cite{yang2019r3det} and CFC-Net \cite{ming2021cfc} that also adopt anchor refinement module. The S$^2$A-Net \cite{han2021align} trained with RIL achieves a mAP of 77.62\%, surpassing the other compared methods.  More visualization of detection results on the DOTA dataset can be found in the appendix.

\begin{table}[t]
	\begin{center}
	\footnotesize
	\renewcommand\arraystretch{1.}
	\setlength{\tabcolsep}{2.5mm}{
		\begin{tabular}{c|cc|l}
			\toprule
			Methods & car  & airplane & $\text{mAP}$ \\ 
			\hline
			YOLOv3  \cite{redmon2018yolov3}                             & 74.63& 89.52 & 82.08     \\ 
			RetinaNet     \cite{lin2017focal}                           & 84.64 &  $\textbf{90.51}$ & 87.57     \\ 
			FR-O \cite{xia2018dota}                 & 86.87 & 89.86 & 88.36 \\ 
			RoI Transformer \cite{ding2019learning}& 87.99 & 89.90 & 88.95    \\
			\hline
			RIDet-Q                               & 88.50 &  89.96 &  89.23     \\ 
			RIDet-O                               & $\textbf{88.88}$ &  90.35 & $\textbf{89.62}$    \\ 
			\bottomrule
	\end{tabular}}
\end{center}
\caption{Detection performance on UCAS-AOD dataset.}
	\label{tab6}
\end{table}

\subsubsection{Results on UCAS-AOD}
The UCAS-AOD dataset contains a large number of densely arranged cars and airplanes. The experimental results on the UCAS-AOD dataset are shown in Tab. \ref{tab6}. RIDet-Q and RIDet-O achieve the mAP of 89.23\% and 89.62\%, respectively.  RIDet outperforms other advanced rotation detectors such as CFC-Net \cite{ming2021cfc} and RoI Transformer \cite{ding2019learning} in our experiments, which proves the superiority of our method.  The visualization results of vehicle detection of RIDet-Q are shown in Fig.~\ref{fig7}. RIL treats the quadrilateral regression as the point matching problem, which signi cantly improves detection performance.

\begin{figure*}[t]
\begin{center}
\includegraphics[width=0.8\linewidth]{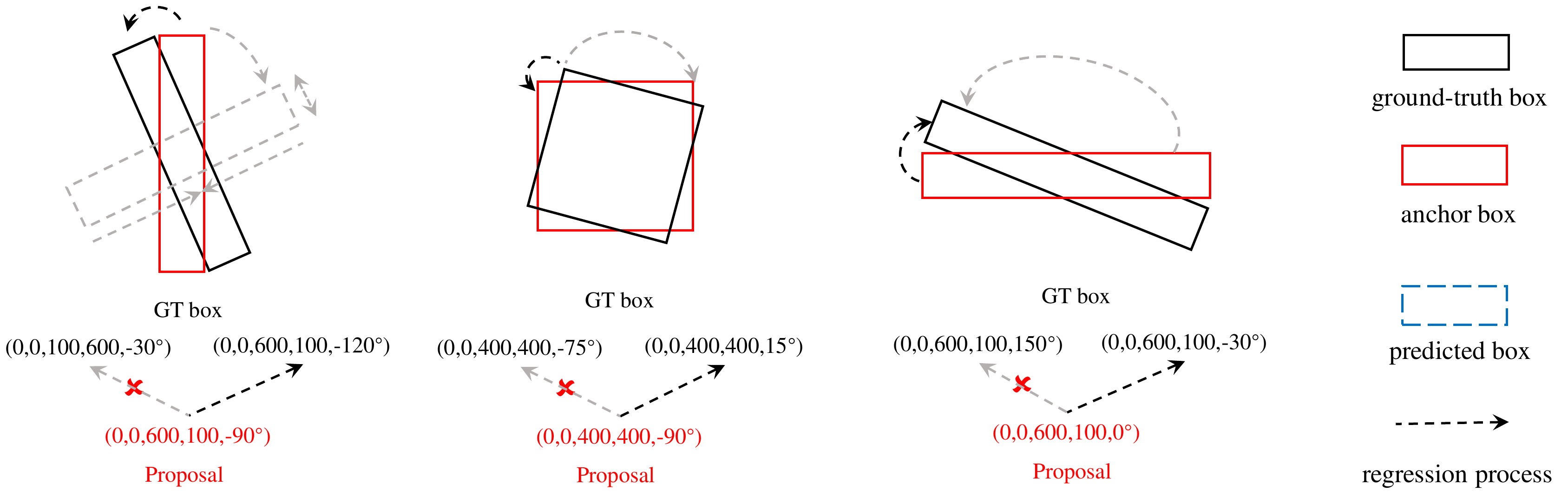}
\end{center}
   \caption{The bounding box regression process of oriented object under QBB (top) and OBB (bottom) characterization. The red box is the GT box, and the blue box is the prediction.}
\label{fig8}
\end{figure*}

 \section{Conclusion}
In this paper, we discuss the suboptimal regression problem caused by the ambiguous representations in oriented object detection. The Representation Invariance Loss (RIL) is proposed to solve the representation ambiguity and improve network convergence to improve detection performance. RIL treats multiple representations as equivalent local minima, and then transforms the bounding box regression task into the optimal matching process between predictions and these local minima. Extensive experiments on multiple datasets proved the superiority of our method.

 \section{Appendix}

\subsection{The Mapping between Object and Representation}

We denote an object as a set $\mathcal{O}$ with one element, and its all possible representations as a set $\mathcal{R}$. Ideally, $f:~\mathcal{R}\rightarrow~\mathcal{O}$ is a bijective function that will not cause ambiguity. However, due to the disorder of the quadrilateral under the QBB and the interchangeability between the width and height under the OBB, this conclusion does not hold. The existence of multiple representations makes $f:~\mathcal{R}\rightarrow~\mathcal{O}$ a surjection. And specific representation can specify the unique object. Therefore, we can use multiple representations as the local minimum for optimization. This conclusion is obviously established in the OBB representation. For QBB representation, this conclusion is only valid when the ground-truth box  is a convex quadrilateral. Fortunately, almost all oriented objects that use quadrilateral labels are convex polygons, such as arbitrary-oriented scene text, remote sensing objects, human faces, and retail products. Therefore, our RIL can effectively use disorder vertices to optimize regression.

\begin{figure}[t]
\begin{center}
\includegraphics[width=0.9\linewidth]{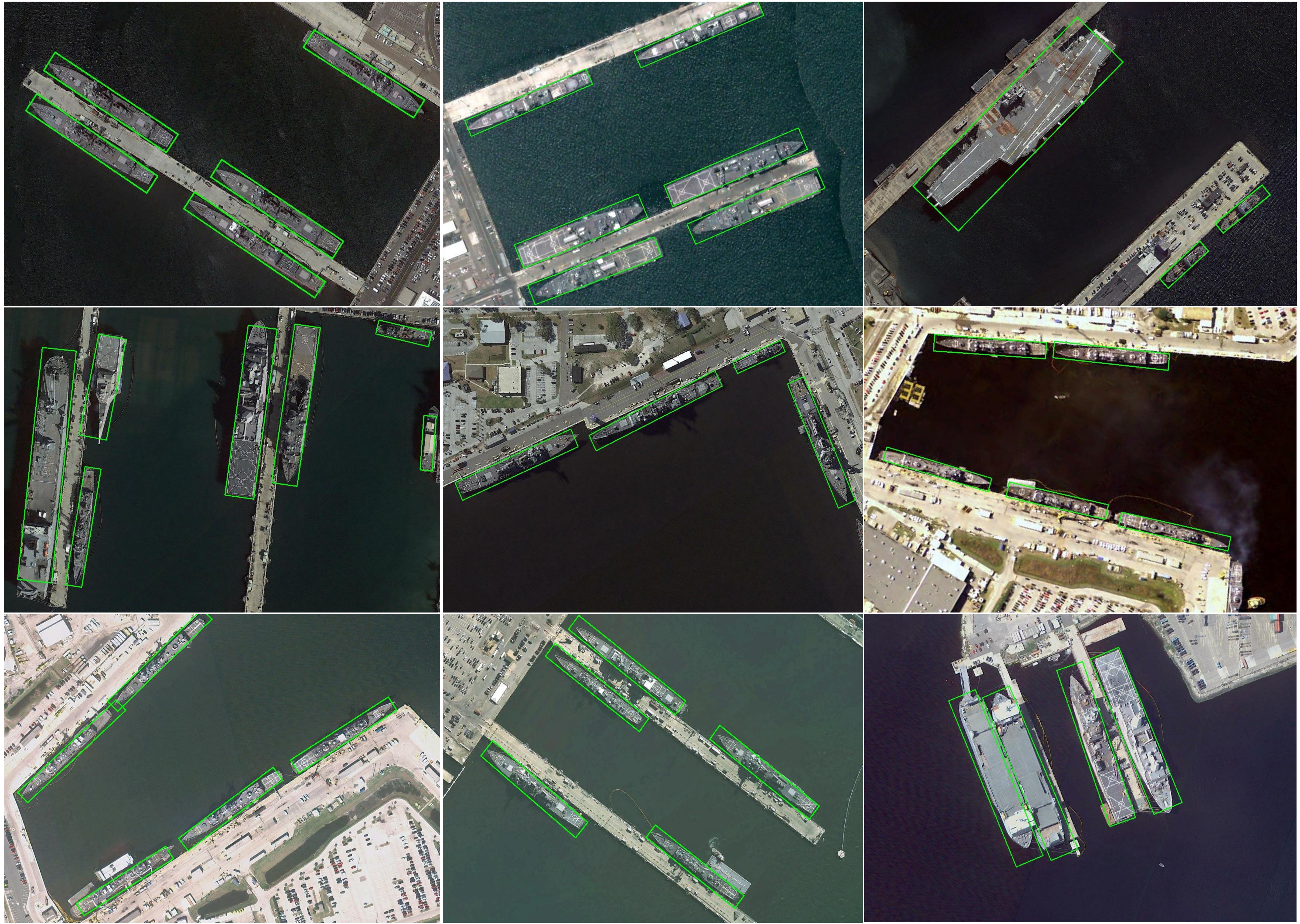}
\end{center}
   \caption{Visualization of detection results on HRSC2016 dataset.}
\label{fig9}
\end{figure}

\begin{figure}[t]
\begin{center}
\includegraphics[width=0.9\linewidth]{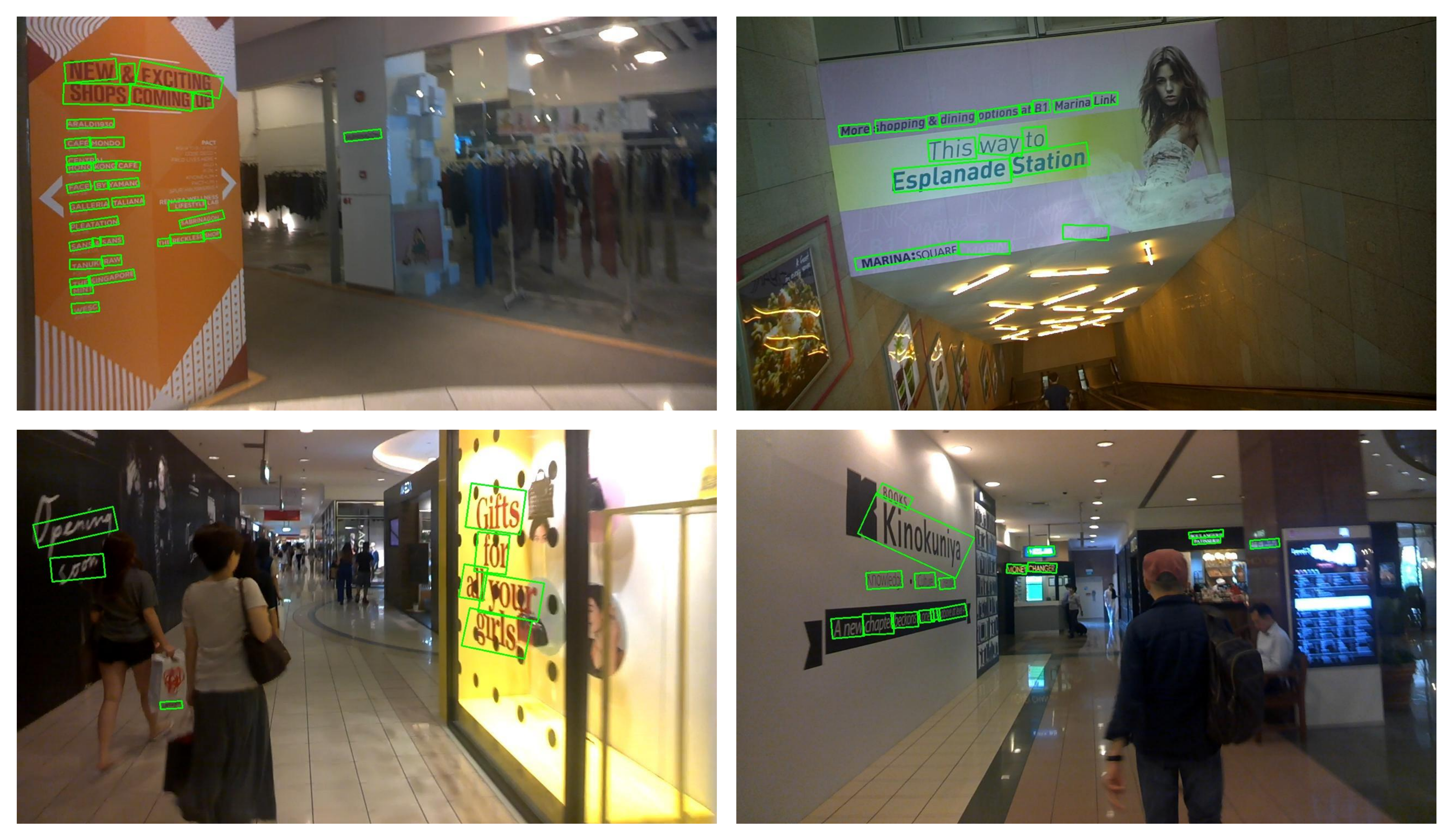}
\end{center}
   \caption{Visualization of detection results on ICDAR2015 dataset.}
\label{fig10}
\end{figure}

\begin{figure*}[t]
\begin{center}
\includegraphics[width=0.9\linewidth]{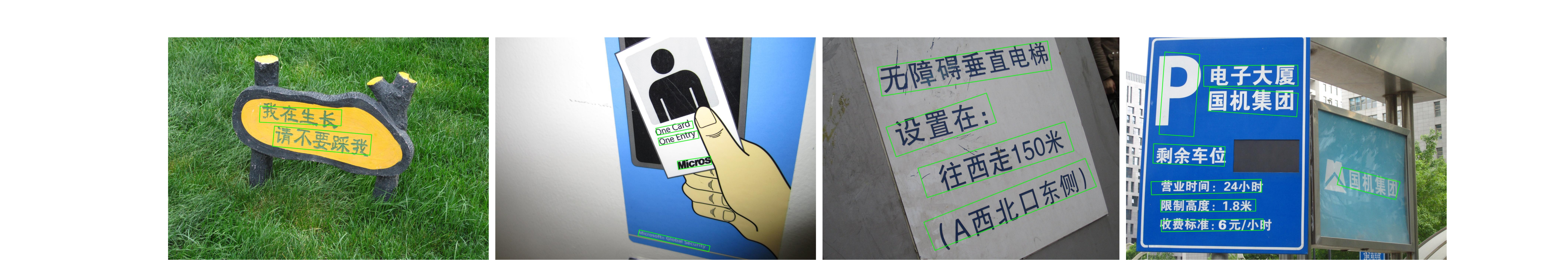}
\end{center}
   \caption{Visualization of detection results on MSRA-TD500 dataset.}
\label{fig11}
\end{figure*}

\begin{figure*}[t]
\begin{center}
\includegraphics[width=0.9\linewidth]{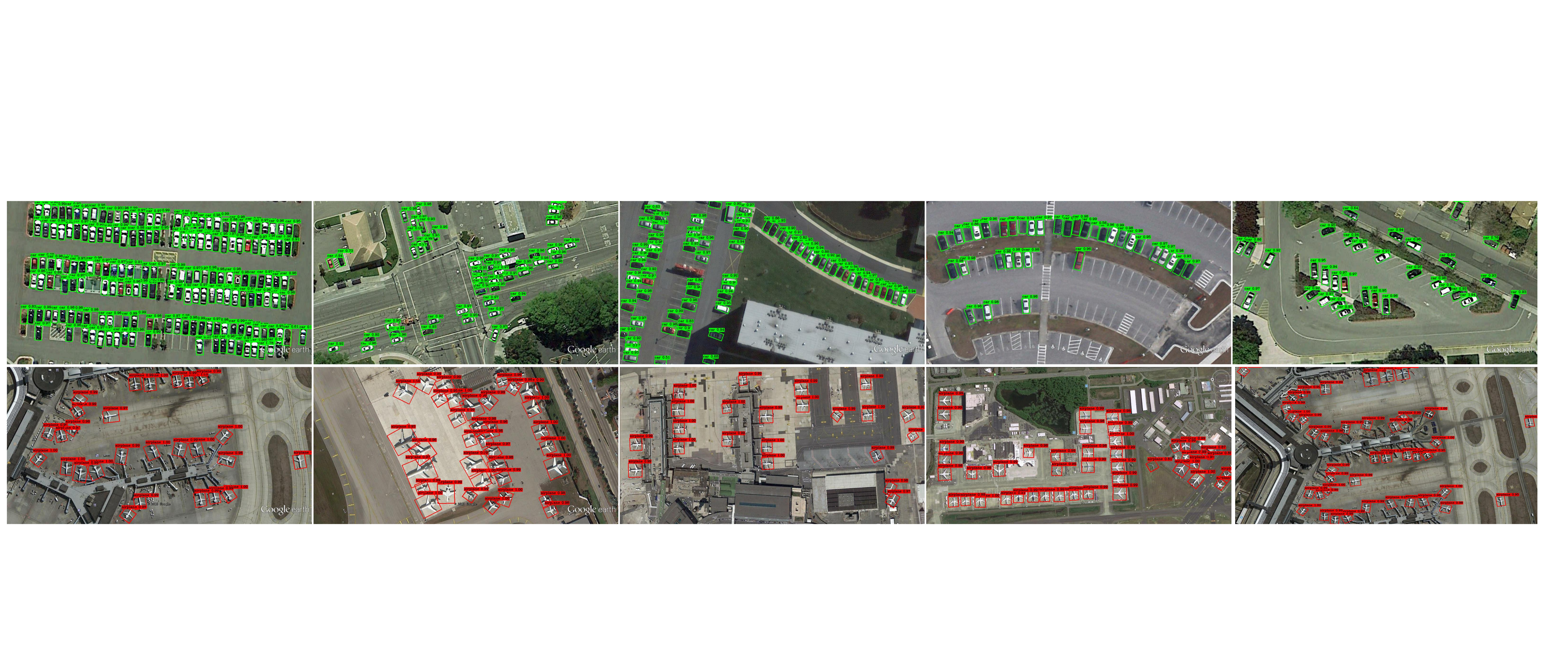}
\end{center}
   \caption{Visualization of detection results on UCAS-AOD dataset}
\label{fig12}
\end{figure*}

\begin{figure*}[h]
\begin{center}
\includegraphics[width=0.9\linewidth]{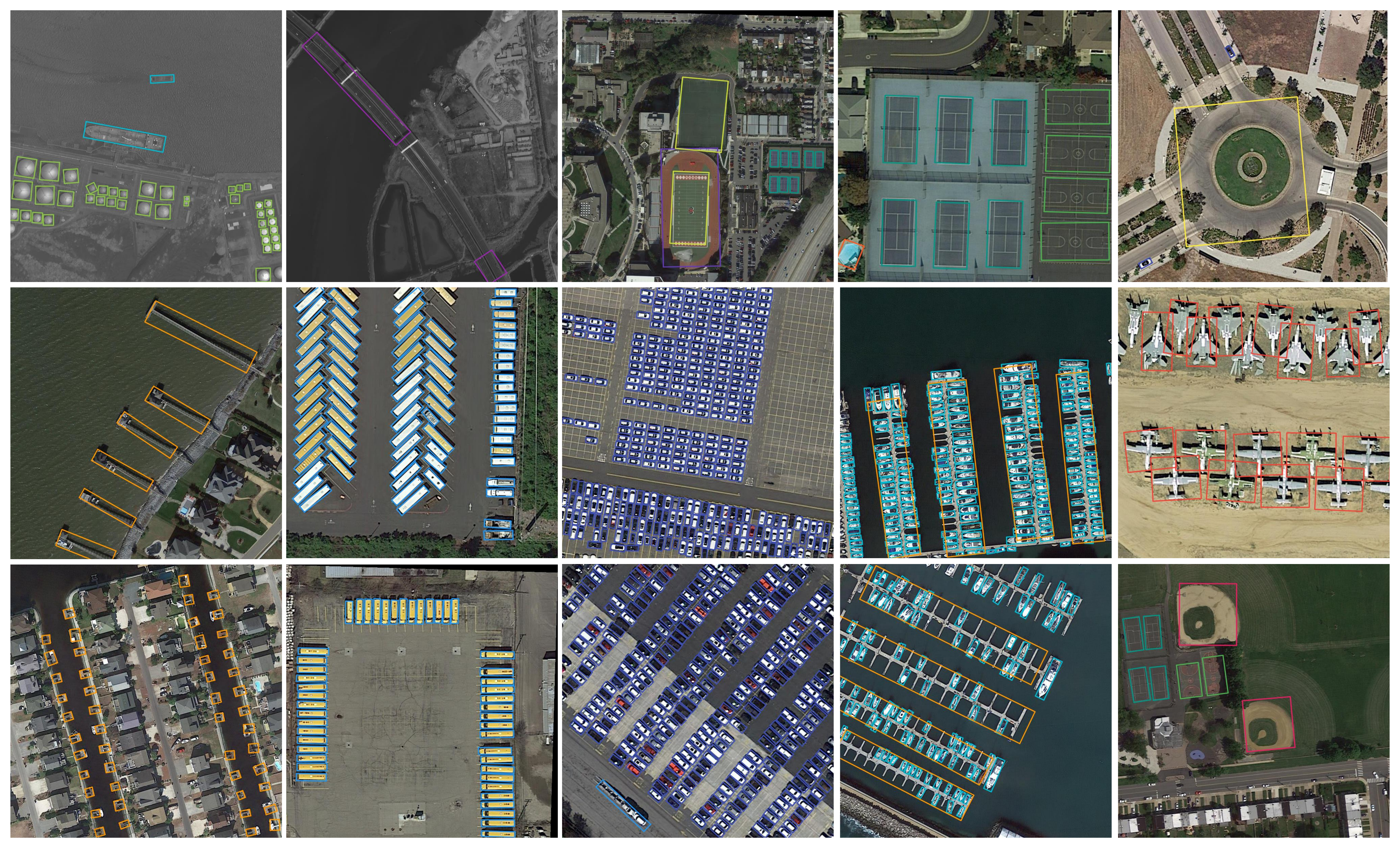}
\end{center}
   \caption{Visualization of detection results on DOTA dataset}
\label{fig13}
\end{figure*}

\subsection{Eliminating Representation Ambiguity via Optimal Matching for OBB}
The matching-based method can effectively eliminate the problems caused by ambiguous representation, which has been introduced in the main body of the paper. Here is a supplementary explanation for the elimination of ambiguity under the OBB representation.

As shown in the left of Fig.~\ref{fig8}, RIL searches for the nearest representation when optimizing the regression loss and fits this minimum value with boundary-free predictions. In this way, the loss function will not increase suddenly but will fit the GT box in the fastest way.

The middle of Fig.~\ref{fig8} shows the solution of RIL in the square-like problem \cite{yang2020dense}. In this case, due to the extreme similarity of width and height, there are two local optima. However, the constraints of definition boundary of the angle hinder free regression of angle, and current detectors cannot be optimized efficiently. RIL can effectively use all redundant representations for optimal matching and resolve this problem.

The illustration on the right in Fig.~\ref{fig8} is similar to the one on the left, so it won’t be elaborated here. The root cause of the ambiguity in the 180$^\circ$ representation is the boundedness of the angle definition, while the 90$^\circ$ representation also suffers from the interchangeability between the edges and the angle. It is worth noting that the normalized rotation mapping loss uses the angular deviation between the two boxes to optimize the orientation and constrain it into $[0,90^\circ]$. Therefore, it can independently solve the representation ambiguity caused by the out-of-bounds definition of angle, such as the situation on the right side of Fig.~\ref{fig8}.

\subsection{Visualization of Detection Results}
We visualized some detection results on different datasets, including three remote sensing data sets: HRSC2016 (shown in Fig.~\ref{fig9}), UCAS-AOD (shown in Fig.~\ref{fig12}), DOTA (shown in Fig.~\ref{fig13}), and two scene text detection datasets: ICDAR2015 (shown in Fig.~\ref{fig10}), MSRA-TD500 (shown in Fig.~\ref{fig11}).

{\small
\bibliographystyle{ieee_fullname}
\bibliography{egbib}

\begin{thebibliography}{10}\itemsep=-1pt

\bibitem{azimi2018towards}
Seyed~Majid Azimi, Eleonora Vig, Reza Bahmanyar, Marco K{\"o}rner, and Peter
  Reinartz.
\newblock Towards multi-class object detection in unconstrained remote sensing
  imagery.
\newblock In {\em Asian Conference on Computer Vision}, pages 150--165.
  Springer, 2018.

\bibitem{ding2019learning}
Jian Ding, Nan Xue, Yang Long, Gui-Song Xia, and Qikai Lu.
\newblock Learning roi transformer for oriented object detection in aerial
  images.
\newblock In {\em Proceedings of the IEEE/CVF Conference on Computer Vision and
  Pattern Recognition}, pages 2849--2858, 2019.

\bibitem{feng2020toso}
Pengming Feng, Youtian Lin, Jian Guan, Guangjun He, Huifeng Shi, and Jonathon
  Chambers.
\newblock Toso: Student’st distribution aided one-stage orientation target
  detection in remote sensing images.
\newblock In {\em ICASSP 2020-2020 IEEE International Conference on Acoustics,
  Speech and Signal Processing (ICASSP)}, pages 4057--4061. IEEE, 2020.

\bibitem{girshick2014rich}
Ross Girshick, Jeff Donahue, Trevor Darrell, and Jitendra Malik.
\newblock Rich feature hierarchies for accurate object detection and semantic
  segmentation.
\newblock In {\em Proceedings of the IEEE conference on computer vision and
  pattern recognition}, pages 580--587, 2014.

\bibitem{han2021align}
J. {Han}, J. {Ding}, J. {Li}, and G.~S. {Xia}.
\newblock Align deep features for oriented object detection.
\newblock {\em IEEE Transactions on Geoscience and Remote Sensing}, pages
  1--11, 2021.

\bibitem{he2017single}
Pan He, Weilin Huang, Tong He, Qile Zhu, Yu Qiao, and Xiaolin Li.
\newblock Single shot text detector with regional attention.
\newblock In {\em Proceedings of the IEEE international conference on computer
  vision}, pages 3047--3055, 2017.

\bibitem{jiang2017r2cnn}
Yingying Jiang, Xiangyu Zhu, Xiaobing Wang, Shuli Yang, Wei Li, Hua Wang, Pei
  Fu, and Zhenbo Luo.
\newblock R2cnn: rotational region cnn for orientation robust scene text
  detection.
\newblock {\em arXiv preprint arXiv:1706.09579}, 2017.

\bibitem{jiao2018densely}
Jiao Jiao, Yue Zhang, Hao Sun, Xue Yang, Xun Gao, Wen Hong, Kun Fu, and Xian
  Sun.
\newblock A densely connected end-to-end neural network for multiscale and
  multiscene sar ship detection.
\newblock {\em IEEE Access}, 6:20881--20892, 2018.

\bibitem{karatzas2015icdar}
Dimosthenis Karatzas, Lluis Gomez-Bigorda, Anguelos Nicolaou, Suman Ghosh,
  Andrew Bagdanov, Masakazu Iwamura, Jiri Matas, Lukas Neumann,
  Vijay~Ramaseshan Chandrasekhar, Shijian Lu, et~al.
\newblock Icdar 2015 competition on robust reading.
\newblock In {\em 2015 13th International Conference on Document Analysis and
  Recognition (ICDAR)}, pages 1156--1160. IEEE, 2015.

\bibitem{kuhn1955hungarian}
Harold~W Kuhn.
\newblock The hungarian method for the assignment problem.
\newblock {\em Naval research logistics quarterly}, 2(1-2):83--97, 1955.

\bibitem{liao2018textboxes++}
Minghui Liao, Baoguang Shi, and Xiang Bai.
\newblock Textboxes++: A single-shot oriented scene text detector.
\newblock {\em IEEE transactions on image processing}, 27(8):3676--3690, 2018.

\bibitem{liao2018rotation}
Minghui Liao, Zhen Zhu, Baoguang Shi, Gui-song Xia, and Xiang Bai.
\newblock Rotation-sensitive regression for oriented scene text detection.
\newblock In {\em Proceedings of the IEEE conference on computer vision and
  pattern recognition}, pages 5909--5918, 2018.

\bibitem{lin2017focal}
Tsung-Yi Lin, Priya Goyal, Ross Girshick, Kaiming He, and Piotr Doll{\'a}r.
\newblock Focal loss for dense object detection.
\newblock In {\em Proceedings of the IEEE international conference on computer
  vision}, pages 2980--2988, 2017.

\bibitem{liu2020efn}
Jin Liu and Haokun Zheng.
\newblock Efn: Field-based object detection for aerial images.
\newblock {\em Remote Sensing}, 12(21):3630, 2020.

\bibitem{liu2016ssd}
Wei Liu, Dragomir Anguelov, Dumitru Erhan, Christian Szegedy, Scott Reed,
  Cheng-Yang Fu, and Alexander~C Berg.
\newblock Ssd: Single shot multibox detector.
\newblock In {\em European conference on computer vision}, pages 21--37.
  Springer, 2016.

\bibitem{liu2019omnidirectional}
Yuliang Liu, Sheng Zhang, Lianwen Jin, Lele Xie, Yaqiang Wu, and Zhepeng Wang.
\newblock Omnidirectional scene text detection with sequential-free box
  discretization.
\newblock {\em IJCAI}, 2019.

\bibitem{liu2017high}
Zikun Liu, Liu Yuan, Lubin Weng, and Yiping Yang.
\newblock A high resolution optical satellite image dataset for ship
  recognition and some new baselines.
\newblock In {\em Proceedings of the International Conference on Pattern
  Recognition Applications and Methods}, volume~2, pages 324--331, 2017.

\bibitem{ma2018arbitrary}
Jianqi Ma, Weiyuan Shao, Hao Ye, Li Wang, Hong Wang, Yingbin Zheng, and
  Xiangyang Xue.
\newblock Arbitrary-oriented scene text detection via rotation proposals.
\newblock {\em IEEE Transactions on Multimedia}, 20(11):3111--3122, 2018.

\bibitem{ming2021cfc}
Qi Ming, Lingjuan Miao, Zhiqiang Zhou, and Yunpeng Dong.
\newblock Cfc-net: A critical feature capturing network for arbitrary-oriented
  object detection in remote-sensing images.
\newblock {\em IEEE Transactions on Geoscience and Remote Sensing}, pages
  1--14, 2021.

\bibitem{ming2021sparse}
Qi Ming, Lingjuan Miao, Zhiqiang Zhou, Junjie Song, and Xue Yang.
\newblock Sparse label assignment for oriented object detection in aerial
  images.
\newblock {\em Remote Sensing}, 13(14):2664, 2021.

\bibitem{ming2020dynamic}
Qi Ming, Zhiqiang Zhou, Lingjuan Miao, Hongwei Zhang, and Linhao Li.
\newblock Dynamic anchor learning for arbitrary-oriented object detection.
\newblock In {\em Proceedings of the AAAI Conference on Artificial
  Intelligence}, volume~35, pages 2355--2363, 2021.

\bibitem{pan2020dynamic}
Xingjia Pan, Yuqiang Ren, Kekai Sheng, Weiming Dong, Haolei Yuan, Xiaowei Guo,
  Chongyang Ma, and Changsheng Xu.
\newblock Dynamic refinement network for oriented and densely packed object
  detection.
\newblock In {\em Proceedings of the IEEE/CVF Conference on Computer Vision and
  Pattern Recognition}, pages 11207--11216, 2020.

\bibitem{paszke2019pytorch}
Adam Paszke, Sam Gross, Francisco Massa, Adam Lerer, James Bradbury, Gregory
  Chanan, Trevor Killeen, Zeming Lin, Natalia Gimelshein, Luca Antiga, et~al.
\newblock Pytorch: An imperative style, high-performance deep learning library.
\newblock {\em arXiv preprint arXiv:1912.01703}, 2019.

\bibitem{qian2019learning}
Wen Qian, Xue Yang, Silong Peng, Yue Guo, and Junchi Yan.
\newblock Learning modulated loss for rotated object detection.
\newblock {\em arXiv preprint arXiv:1911.08299}, 2019.

\bibitem{redmon2016you}
Joseph Redmon, Santosh Divvala, Ross Girshick, and Ali Farhadi.
\newblock You only look once: Unified, real-time object detection.
\newblock In {\em Proceedings of the IEEE conference on computer vision and
  pattern recognition}, pages 779--788, 2016.

\bibitem{redmon2018yolov3}
Joseph Redmon and Ali Farhadi.
\newblock Yolov3: An incremental improvement.
\newblock {\em arXiv preprint arXiv:1804.02767}, 2018.

\bibitem{ren2016faster}
Shaoqing Ren, Kaiming He, Ross Girshick, and Jian Sun.
\newblock Faster r-cnn: towards real-time object detection with region proposal
  networks.
\newblock {\em IEEE transactions on pattern analysis and machine intelligence},
  39(6):1137--1149, 2016.

\bibitem{shi2018real}
Xuepeng Shi, Shiguang Shan, Meina Kan, Shuzhe Wu, and Xilin Chen.
\newblock Real-time rotation-invariant face detection with progressive
  calibration networks.
\newblock In {\em Proceedings of the IEEE Conference on Computer Vision and
  Pattern Recognition}, pages 2295--2303, 2018.

\bibitem{song2020learning}
Qing Song, Fan Yang, Lu Yang, Chun Liu, Mengjie Hu, and Lurui Xia.
\newblock Learning point-guided localization for detection in remote sensing
  images.
\newblock {\em IEEE Journal of Selected Topics in Applied Earth Observations
  and Remote Sensing}, 2020.

\bibitem{wei2020oriented}
Haoran Wei, Yue Zhang, Zhonghan Chang, Hao Li, Hongqi Wang, and Xian Sun.
\newblock Oriented objects as pairs of middle lines.
\newblock {\em ISPRS Journal of Photogrammetry and Remote Sensing},
  169:268--279, 2020.

\bibitem{xia2018dota}
Gui-Song Xia, Xiang Bai, Jian Ding, Zhen Zhu, Serge Belongie, Jiebo Luo, Mihai
  Datcu, Marcello Pelillo, and Liangpei Zhang.
\newblock Dota: A large-scale dataset for object detection in aerial images.
\newblock In {\em Proceedings of the IEEE Conference on Computer Vision and
  Pattern Recognition}, pages 3974--3983, 2018.

\bibitem{xu2020gliding}
Yongchao Xu, Mingtao Fu, Qimeng Wang, Yukang Wang, Kai Chen, Gui-Song Xia, and
  Xiang Bai.
\newblock Gliding vertex on the horizontal bounding box for multi-oriented
  object detection.
\newblock {\em IEEE transactions on pattern analysis and machine intelligence},
  2020.

\bibitem{yang2020dense}
Xue Yang, Liping Hou, Yue Zhou, Wentao Wang, and Junchi Yan.
\newblock Dense label encoding for boundary discontinuity free rotation
  detection.
\newblock {\em arXiv preprint arXiv:2011.09670}, 2020.

\bibitem{yang2019r3det}
Xue Yang, Qingqing Liu, Junchi Yan, Ang Li, Zhiqiang Zhang, and Gang Yu.
\newblock R3det: Refined single-stage detector with feature refinement for
  rotating object.
\newblock {\em arXiv preprint arXiv:1908.05612}, 2019.

\bibitem{yang2018automatic}
Xue Yang, Hao Sun, Kun Fu, Jirui Yang, Xian Sun, Menglong Yan, and Zhi Guo.
\newblock Automatic ship detection in remote sensing images from google earth
  of complex scenes based on multiscale rotation dense feature pyramid
  networks.
\newblock {\em Remote Sensing}, 10(1):132, 2018.

\bibitem{yang2018position}
Xue Yang, Hao Sun, Xian Sun, Menglong Yan, Zhi Guo, and Kun Fu.
\newblock Position detection and direction prediction for arbitrary-oriented
  ships via multitask rotation region convolutional neural network.
\newblock {\em IEEE Access}, 6:50839--50849, 2018.

\bibitem{yang2020arbitrary}
Xue Yang and Junchi Yan.
\newblock Arbitrary-oriented object detection with circular smooth label.
\newblock In {\em European Conference on Computer Vision}, pages 677--694.
  Springer, 2020.

\bibitem{yang2021rethinking}
Xue Yang, Junchi Yan, Qi Ming, Wentao Wang, Xiaopeng Zhang, and Qi Tian.
\newblock Rethinking rotated object detection with gaussian wasserstein
  distance loss.
\newblock {\em arXiv preprint arXiv:2101.11952}, 2021.

\bibitem{yang2020scrdet++}
Xue Yang, Junchi Yan, Xiaokang Yang, Jin Tang, Wenglong Liao, and Tao He.
\newblock Scrdet++: Detecting small, cluttered and rotated objects via
  instance-level feature denoising and rotation loss smoothing.
\newblock {\em arXiv preprint arXiv:2004.13316}, 2020.

\bibitem{yang2019scrdet}
Xue Yang, Jirui Yang, Junchi Yan, Yue Zhang, Tengfei Zhang, Zhi Guo, Xian Sun,
  and Kun Fu.
\newblock Scrdet: Towards more robust detection for small, cluttered and
  rotated objects.
\newblock In {\em Proceedings of the IEEE/CVF International Conference on
  Computer Vision}, pages 8232--8241, 2019.

\bibitem{yao2012detecting}
Cong Yao, Xiang Bai, Wenyu Liu, Yi Ma, and Zhuowen Tu.
\newblock Detecting texts of arbitrary orientations in natural images.
\newblock In {\em 2012 IEEE conference on computer vision and pattern
  recognition}, pages 1083--1090. IEEE, 2012.

\bibitem{yi2020oriented}
Jingru Yi, Pengxiang Wu, Bo Liu, Qiaoying Huang, Hui Qu, and Dimitris Metaxas.
\newblock Oriented object detection in aerial images with box boundary-aware
  vectors.
\newblock In {\em Proceedings of the IEEE/CVF Winter Conference on Applications
  of Computer Vision}, pages 2150--2159, 2020.

\bibitem{zhang2019cad}
Gongjie Zhang, Shijian Lu, and Wei Zhang.
\newblock Cad-net: A context-aware detection network for objects in remote
  sensing imagery.
\newblock {\em IEEE Transactions on Geoscience and Remote Sensing},
  57(12):10015--10024, 2019.

\bibitem{zhang2020bridging}
Shifeng Zhang, Cheng Chi, Yongqiang Yao, Zhen Lei, and Stan~Z Li.
\newblock Bridging the gap between anchor-based and anchor-free detection via
  adaptive training sample selection.
\newblock In {\em Proceedings of the IEEE/CVF Conference on Computer Vision and
  Pattern Recognition}, pages 9759--9768, 2020.

\bibitem{zhang2018toward}
Zenghui Zhang, Weiwei Guo, Shengnan Zhu, and Wenxian Yu.
\newblock Toward arbitrary-oriented ship detection with rotated region proposal
  and discrimination networks.
\newblock {\em IEEE Geoscience and Remote Sensing Letters}, 15(11):1745--1749,
  2018.

\bibitem{zheng2020rotation}
Yu Zheng, Danyang Zhang, Sinan Xie, Jiwen Lu, and Jie Zhou.
\newblock Rotation-robust intersection over union for 3d object detection.
\newblock In {\em European Conference on Computer Vision}, pages 464--480.
  Springer, 2020.

\bibitem{zhou2020objects}
Lin Zhou, Haoran Wei, Hao Li, Wenzhe Zhao, Yi Zhang, and Yue Zhang.
\newblock Objects detection for remote sensing images based on polar
  coordinates.
\newblock {\em arXiv preprint arXiv:2001.02988}, 2020.

\bibitem{zhu2015orientation}
Haigang Zhu, Xiaogang Chen, Weiqun Dai, Kun Fu, Qixiang Ye, and Jianbin Jiao.
\newblock Orientation robust object detection in aerial images using deep
  convolutional neural network.
\newblock In {\em 2015 IEEE International Conference on Image Processing
  (ICIP)}, pages 3735--3739. IEEE, 2015.

\end{thebibliography}
}

\end{document}